\documentclass[10pt,twocolumn,letterpaper]{article}

\usepackage[pagenumbers]{cvpr} % To force page numbers, e.g. for an arXiv version

\usepackage{graphicx}
\usepackage{amsmath}
\usepackage{amssymb}
\usepackage{booktabs}
\usepackage{amssymb}
\usepackage{multirow}
\usepackage{multicol}
\usepackage{bm}
\usepackage{colortbl}
\usepackage{wrapfig}
\usepackage{pifont}
 
\makeatletter
\newcommand{\thickhline}{%
    \noalign {\ifnum 0=`}\fi \hrule height 1.2pt
    \futurelet \reserved@a \@xhline
}
\definecolor{mygray}{gray}{.9}
\newcommand\ourmethod{DAT}
\newcommand\omlong{Deformable Attention Transformer}
\usepackage[pagebackref,breaklinks,colorlinks]{hyperref}

\usepackage[capitalize]{cleveref}
\crefname{section}{Sec.}{Secs.}
\Crefname{section}{Section}{Sections}
\Crefname{table}{Table}{Tables}
\crefname{table}{Tab.}{Tabs.}

\def\cvprPaperID{7365} % *** Enter the CVPR Paper ID here
\def\confName{CVPR}
\def\confYear{2022}

\begin{document}
% \title{DEFEAT:DEFormablE Attention Transformer}
%\title{PVD: Pyramid Vision Deformable Transformer for Dense Prediction}
%\title{Hierarchical Vision Transformer with Deformable Attention}
% \title{Pyramid Vision Deformer: Hierarchical Vision Transformer with Deformable Attention}
%Multi-scale deformable vision transformer for dense prediction
\title{Vision Transformer with Deformable Attention}
\author{
  Zhuofan Xia$^{1}$\thanks{Equal contribution.}\qquad
  Xuran Pan$^{1}$\footnotemark[1]\qquad
  Shiji Song$^{1}$\qquad
  Li Erran Li$^{2}$\qquad
  Gao Huang$^{1,3}$\thanks{Corresponding author.}\\
    \normalsize{$^{1}$Department of Automation, BNRist, Tsinghua University} \\
    \normalsize{$^{2}$AWS AI, Amazon \quad $^{3}$Beijing Academy of Artificial Intelligence}
%   \texttt{\small \{xzf20, pxr18\}@mails.tsinghua.edu.cn}, \texttt{\small{} erranlli@gmail.com},\\
%   \texttt{\small\{shijis, gaohuang\}@tsinghua.edu.cn}
}
\maketitle

%%%%%%%%% ABSTRACT
\begin{abstract}

%   With the recent prosperity of Vision Transformers in 
Transformers have recently shown superior performances on various vision tasks. The large, sometimes even global, receptive field endows Transformer models with higher representation power over their CNN counterparts. Nevertheless, simply enlarging receptive field also gives rise to several concerns. 
%^(1) Features can be influenced by irrelevant parts, which are beyond the region of interests. (2) 
On the one hand, using dense attention e.g., in ViT, leads to excessive memory and computational cost, and features can be influenced by irrelevant parts which are beyond the region of interests. On the other hand, the sparse attention adopted in PVT or Swin Transformer is data agnostic and may limit the ability to model long range relations.
% suffers from the information loss where
% the fixed pattern in the query-key-value computation leads to limited capacity on modeling geometric transformations. 
% the data-agnostic attention pattern may drop relevant keys and values.
To mitigate these issues, we propose a novel deformable self-attention module, where the positions of key and value pairs in self-attention are selected in a data-dependent way. This flexible scheme enables the self-attention module to focus on relevant regions and capture more informative features. On this basis, we present \textbf{\omlong{}}, a general backbone model with deformable attention for both image classification and dense prediction tasks. Extensive experiments show that our models achieve consistently improved results on comprehensive benchmarks. Code is available at \url{https://github.com/LeapLabTHU/DAT}.

% The key challenge of adapting transformers to vision tasks is effective attention mechanisms. Vanilla self-attention has a computation cost quadratic with respect to input visual token length and results in slow convergence as each token attends to irrelevant information. Windowed attention trades off a linear computation cost with restricted information to attend. In this paper, we propose a simple and efficient deformable self-attention which can learn a few groups of query agnostic offsets to shift keys and values to important regions. This flexible scheme enables the self-attention module to focus on relevant regions and capture more informative features. On this basis, we present \textbf{\omlong{}}, a general backbone model with deformable attention for both image classification and dense prediction tasks. Extensive experiments show that our models achieve consistently improved results on comprehensive benchmarks.

\end{abstract}

% \begin{figure}
% 	\centering
% 	\subfloat[ViT \cite{ViT16x16}]{\label{fig:fig1_1}\includegraphics[width=0.43\linewidth]{figures/fig1_1.pdf}}\quad
% 	\subfloat[Swin Transformer \cite{Swin}]{\label{fig:fig1_2}\includegraphics[width=0.43\linewidth]{figures/fig1_2.pdf}} \\
% 	\subfloat[DCN \cite{DCNv1}]{\label{fig:fig1_3}\includegraphics[width=0.43\linewidth]{figures/fig1_3.pdf}}\quad
% 	\subfloat[\ourmethod{} (ours)]{\label{fig:fig1_4}\includegraphics[width=0.43\linewidth]{figures/fig1_4.pdf}}
% 	\caption{
% % 	Comparison of attention mechanisms between our method and popular Vision Transformers. 
% 	%The red star denotes the query token, the orange mask with solid line boundaries denotes the region to which the query attends. In Figure \ref{fig:fig1_4}, the pink dots are the sampled key-value points to be performed attention on.
% % 	The stars denote the query tokens, the orange mask with solid line boundaries denotes the region to which the query attends. In Figure \ref{fig:fig1_4}, the purple dots are the sampled key-value points for query tokens to attend to.
% 	Comparison of DAT with other Vision Transformer models and DCN in CNN model. The star in red and blue denote the different queries, and mask with solid line boundaries denotes the region to which the query attends. In a data-agnostic way: (a) ViT adopts full attention for all queries. (b) Swin Transformer uses partitioned window attention. In a data-dependent way: (c) DCN learns different offsets for each query. (d) DAT learns shared offsets for all queries.  
% 	}
% \vskip -0.1in
% \end{figure}

\begin{figure}
    \centering
    \includegraphics[width=0.95\linewidth]{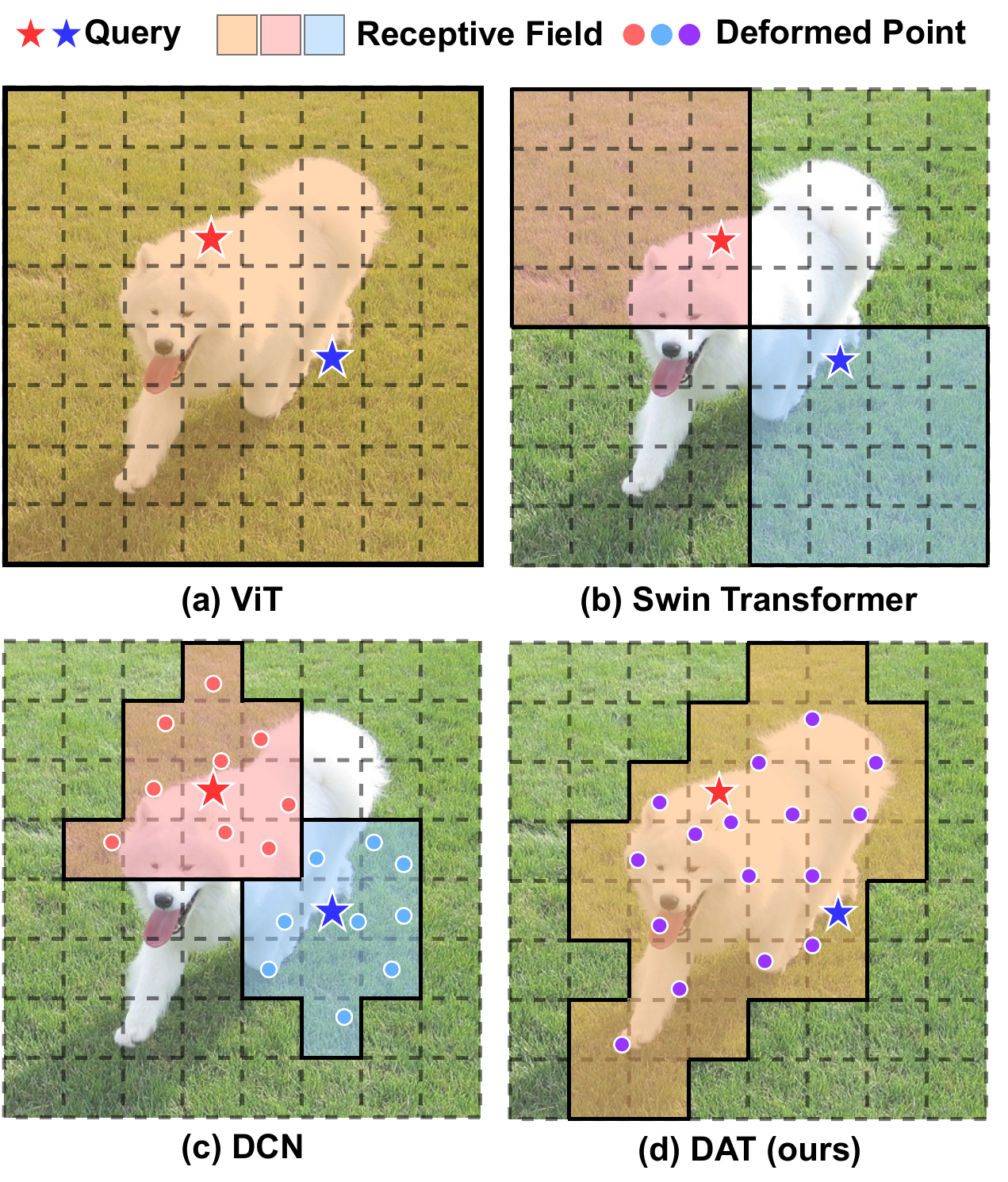}
    \caption{Comparison of DAT with other Vision Transformer models and DCN in CNN model. The red star and the blue star denote the different queries, and masks with solid line boundaries denote the regions to which the queries attend. In a data-agnostic way: (a) ViT \cite{ViT16x16} adopts full attention for all queries. (b) Swin Transformer \cite{Swin} uses partitioned window attention. In a data-dependent way: (c) DCN \cite{DCNv1} learns different deformed points for each query. (d) DAT learns shared deformed points for all queries.}
    \label{fig:fig1}
    \vskip -0.1in
\end{figure}
\vskip -0.1in

\section{Introduction}

% XZF Story line:
% Strengths & draw backs of ViT -> excessive attention -> PVT / Swin drop information data agnostic
% -> selectively drop information, requirements of an ideal target -> CNN also meet this problem -> DCN -> 
% difficulty in memory, D-DETR/DPT/PSVIT -> DAT design -> justification of shared kvs

Transformer \cite{attention} is originally introduced to solve natural language processing tasks. It has recently shown great potential in the field of computer vision \cite{ViT16x16, Swin, PVT}. The pioneer work, Vision Transformer \cite{ViT16x16} (ViT), stacks multiple Transformer blocks to process non-overlapping image patch (\textit{i.e.} visual token) sequences, leading to a convolution-free model for image classification. Compared to their CNN counterparts \cite{resnet,densenet}, Transformer-based models have larger receptive fields and excel at modeling long-range dependencies, which are proved to achieve superior performance in the regime of a large amount of training data and model parameters. However, the superfluous attention in visual recognition is a double-edged sword, and has multiple drawbacks.
%erran: with the excessive attendable keys that trap the optimization and hunger for massive data.
%In specific, 
Specifically,
the excessive number of keys to attend per query patch yields high computational cost and slow convergence, and increases the risk of overfitting.
% Even though the data efficiency is improved by several subsequent works \cite{deit,t2t,lvvit} achieving promising results on ImageNet \cite{in1k}, the excessive attention with quadratic computational complexity still 
%erran: hedges 
% prevents
% Vision Transformer from adapting to various dense prediction tasks, e.g., object detection and semantic segmentation.
% , in which the feature map resolution is much higher and multi-scale feature maps are required.
% INSTEAD OF SAYING SINGLE SCALE ARCH. WE ARGUE THE DOUBLE EDGE SWORD OF LARGE ATTENTION RECPTIVE FIELD.
% the single-scale architecture of ViT and the quadratic computational complexity of multi-head self-attention pose limitations on dense prediction tasks, e.g., object detection and semantic segmentation, in which the input resolution are much higher and features from multiple resolutions are required. 

In order to avoid excessive attention computation, existing works \cite{PVT,Swin,vil,cswin,focal,rgvit}  have 
% adopt pyramid model structures for dense prediction tasks and design various attention patterns to 
%erran: drop the redundant information in attention with priors. 
% reduce the amount of information in attention computation.
% drop the redundant information in attention with priors.
leveraged carefully designed efficient attention patterns to reduce the computation complexity. As two representative approaches among them, Swin Transformer \cite{Swin} adopts window-based local attention to restrict attention in local windows,
% and shifts windows in consecutive layers to achieve information fusion across different windows, 
while Pyramid Vision Transformer (PVT) \cite{PVT}  downsamples the key and value feature maps to save computation. Though effective, the hand-crafted attention patterns are data-agnostic and may not be optimal. It is likely that relevant keys/values are dropped, while less important ones are still kept.

Ideally, one would expect that the candidate key/value set for a given query is flexible and has the ability to adapt to each individual input, such that the issues in hand-crafted sparse attention patterns can be alleviated. In fact, in the literature of CNNs, learning a deformable receptive field for the convolution filters has been shown effective in selectively attending to more informative regions on a data-dependent basis \cite{DCNv1}. The most notable work, Deformable Convolution Networks \cite{DCNv1}, has yielded impressive results on many challenging vision tasks. This motivates us to explore a deformable attention pattern in Vision Transformers. However, a naive implementation of this idea leads to an unreasonably high memory/computation complexity: the overhead introduced by the deformable offsets is quadratic \emph{w.r.t} the number of patches. As a consequence, although some recent work \cite{DeformableDETR, dpt, psvit} have investigated the idea of deformable 
%attention
mechanism in Transformers
, none of them have treated it as a basic building block for constructing a powerful backbone network like the DCN, due to the high computational cost. Instead, their deformable mechanism is either adopted in the detection head \cite{DeformableDETR}, or used as a pre-processing layer to sample patches for the subsequent backbone network \cite{dpt}. 

In this paper, we present a simple and efficient deformable self-attention module, equipped with which a powerful pyramid backbone, named \textit{\omlong{}} (DAT), is constructed for image classification and various dense prediction tasks. 
% Different from CNNs, Vision Transformer has larger receptive field to model longer dependencies, therefore we propose to learn a few groups of query agnostic offsets to shift keys and values to important regions to extract better features, based on the observed in \cite{gcnet,deepvit} that global attention usually results in the almost same attention maps for different queries.
Different from DCN that learns different offsets for different pixels in the whole feature map, we propose to learn a few groups of \textbf{query agnostic} offsets to shift keys and values to important regions (as illustrated in Figure \ref{fig:fig1}(d)), based on the observation in \cite{gcnet,deepvit} that global attention usually results in the almost same attention patterns for different queries. This design both holds a linear space complexity and introduces a deformable attention pattern to Transformer backbones. Specifically, for each attention module, reference points are first generated as uniform grids, which are the same across the input data. Then, an offset network takes as input the query features and generates the corresponding offsets for all the reference points. In this way, the candidate keys/values are shifted towards important regions, thus augmenting the original self-attention module with higher flexibility and efficiency to capture more informative features.
%erran:
%A general pyramid backbone equipped with our deformable attention module, dubbed \textit{\omlong{}}, is also presented to build a powerful Vision Transformer which can adapt to both image classification and dense prediction tasks. 
% We further present a general and powerful pyramid backbone equipped with our deformable attention module, dubbed \textit{\omlong{}}, which can be adapted to both image classification and various dense prediction tasks. 

% The first DATT backbone
To summarize, our contributions are as follows: we propose the first deformable self-attention backbone for visual recognition, where the data-dependent attention pattern endows higher flexibility and efficiency. 
%Extensive experiments on ImageNet \cite{in1k}, ADE20K \cite{ade20k} and COCO \cite{coco} demonstrate that our model outperforms competitive baselines consistently, including $0.7$ improvements on image classification, $1.2$ improvements on semantic segmentation, $1.0$ improvements on object detection for both box AP and mask AP. The advantages on small and large objects are more distinct with $2.7$ improvements.
Extensive experiments on ImageNet \cite{in1k}, ADE20K \cite{ade20k} and COCO \cite{coco} demonstrate that our model outperforms competitive baselines including Swin Transformer consistently, by a margin of 0.7 on the top-1 accuracy of image classification, 1.2 on the mIoU of semantic segmentation, 1.1 on object detection for both box AP and mask AP. The advantages on small and large objects are more distinct with a margin of 2.1.

% (2) We present a carefully designed model architecture, which incorporates with deformable self-attnetion and shows superior performances over competitive baselines on various vision benchmarks. 
% clean idea, backbone, experiments
\section{Related Work}

\textbf{Transformer vision backbone.}
Since the introduction of ViT~\cite{ViT16x16}, improvements~\cite{PVT,Swin,vil,cswin,focal,rgvit,acmix} have focused on learning multi-scale features for dense prediction tasks and efficient attention mechanisms. These attention mechanisms include windowed attention~\cite{Swin, cswin}, global tokens~\cite{rgvit, vparser, perceiver}, focal attention~\cite{focal} and dynamic token sizes~\cite{dvt}.
% Different from these, we propose a novel deformable attention mechanism and demonstrate its effectiveness when applied to vision backbones.
%erran: More recently, researches begin to explore the possibility of introducing convolution-based approaches into Vistion Transformer models. 
More recently, convolution-based approaches have been introduced into Vision Transformer models. 
Among which exist researches focusing on complementing transformer models with convolution operations to introduce additional inductive biases. CvT~\cite{cvt} adopts convolution in the tokenization process and utilizes stride convolution to reduce the computation complexity of self-attention. ViT with convolutional stem~\cite{early} proposes to add convolutions at the early stage to achieve stabler training. CSwin Transformer~\cite{cswin} adopts a convolution-based positional encoding technique and shows improvements on downstream tasks. 
%\ourmethod{} focuses on efficient deformable attention. 
Many of these convolution-based techniques can potentially be applied on top of \ourmethod{} for further performance improvements.

%\cite{DeformableDETR,dpt,psvit}
%sparse attention.

\textbf{Deformable CNN and attention.}
%erran: Deformable mechanism is also one of the convolution techniques that show potential applications in Vision Transformer. There are several works \cite{DeformableDETR,dpt,psvit} attempting to ignite the light of deformable mechanism. Deformable DETR \cite{DeformableDETR} tries to tackle this challenge by selecting only very few keys for each query on the top of a CNN backbone, yet it is infeasible to adopt it to a visual backbone for feature extraction due to the lack of enough keys for better representations. DPT \cite{dpt} and PS-ViT \cite{psvit} builds deformable modules to refine visual tokens. Specifically, DPT propose a deformable patch embedding to refine patches across stages and PS-ViT introduce spatial sampling module before a ViT backbone to improve visual tokens. None of them incorporate deformbale attention into backbones.
Deformable convolution~\cite{DCNv1, DCNv2} is a powerful mechanism to attend to flexible spatial locations conditioned on input data. Recently it has been applied to Vision Transformers~\cite{DeformableDETR,dpt,psvit}. Deformable DETR~\cite{DeformableDETR} improves the convergence of DETR~\cite{detr} by selecting a small number of keys for each query on the top of a CNN backbone. Its deformable attention is not suited to a visual backbone for feature extraction as the lack of keys restricts representation power. Furthermore, the attention in Deformable DETR comes from simply learned linear projections and keys are not shared among query tokens. DPT~\cite{dpt} and PS-ViT \cite{psvit} builds deformable modules to refine visual tokens. Specifically, DPT proposes a deformable patch embedding to refine patches across stages and PS-ViT introduces a spatial sampling module before a ViT backbone to improve visual tokens. None of them incorporate deformable attention into vision backbones. In contrast, our deformable attention takes a powerful and yet simple design to learn a set of global keys shared among visual tokens, and can be adopted as a general backbone for various vision tasks. Our method can also be viewed as a spatial adaptive mechanism, which has been proved effective in various works~\cite{gfnet,sarnet}.

\begin{figure*}
    \centering
    \includegraphics[width=\linewidth]{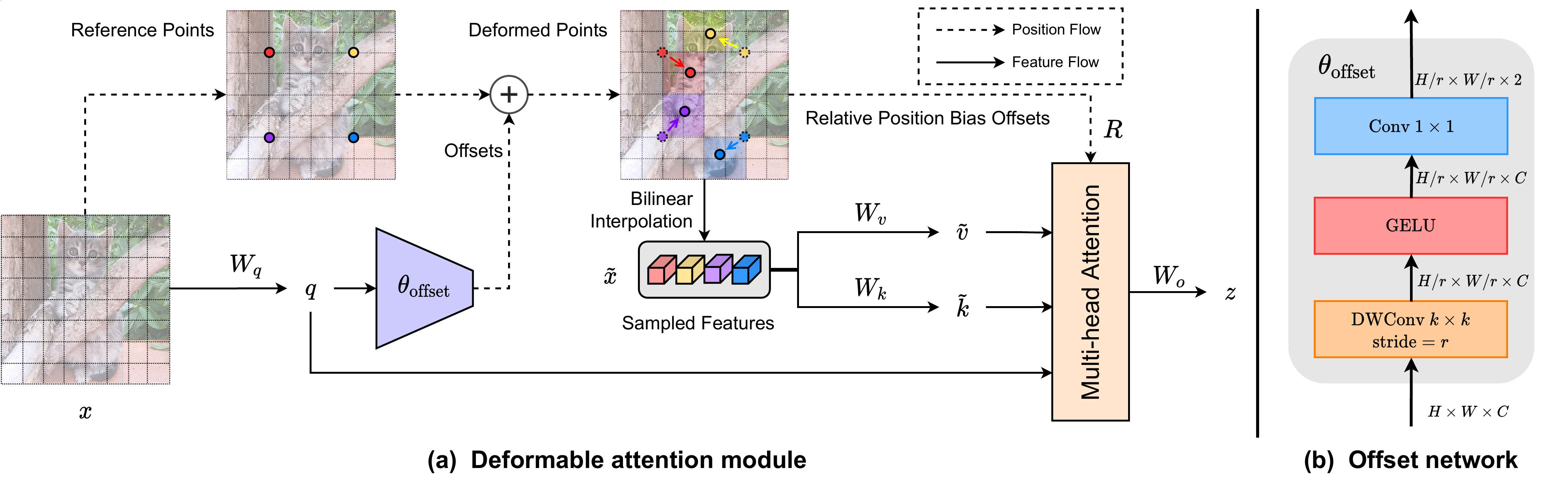}
    \caption{An illustration of our deformable attention mechanism. (a) presents the information flow of deformable attention. In the left part, a group of reference points is placed uniformly on the feature map, whose offsets are learned from the queries by the offset network. Then the deformed keys and values are projected from the sampled features according to the deformed points, as shown in the right part. Relative position bias is also computed by the deformed points, enhancing the multi-head attention which outputs the transformed features. We show only 4 reference points for a clear presentation, there are many more points in real implementation \textit{de facto}. (b) reveals the detailed structure of the offset generation network, marked with sizes of feature maps.}
    \label{fig:fig2}
\end{figure*}
\section{Deformable Attention Transformer}

\subsection{Preliminaries}

We first revisit the attention mechanism in recent Vision Transformers. Taking a flattened feature map $x\in\mathbb{R}^{N\times{}C}$ as the input, a multi-head self-attention (MHSA) block with $M$ heads is formulated as 
\begin{align}
    &q=xW_q,\ k=xW_k,\ v=xW_v, \label{eq:proj} \\
    &z^{(m)}=\sigma(q^{(m)}k^{(m)\top}/\sqrt{d})v^{(m)}, m\!=\!1,\ldots,M, \label{eq:attn} \\
    &z=\text{Concat}\left(z^{(1)},\ldots,z^{(M)}\right)W_o, \label{eq:MHSA}
\end{align}
% where $\sigma(\cdot)$ denotes the softmax function,  $z^{(m)}$ denotes the embeddings output from the $m$-th attention head, $q^{(m)}, k^{(m)}, v^{(m)}\in\mathbb{R}^{N\times{}C}$ denote query, key and value embeddings respectively, $W_q,W_k,W_v,W_o\in\mathbb{R}^{C\times{}C}$ are the projection matrices of this block, $d\!=\!C/M$ is the dimension of each head. To build up a Transformer block, an MLP block with two linear transformations and a GELU activation is usually adopted to provide nonlinearity.
where $\sigma(\cdot)$ denotes the softmax function, and $d\!=\!C/M$ is the dimension of each head. $z^{(m)}$ denotes the embedding output from the $m$-th attention head, $q^{(m)}, k^{(m)}, v^{(m)}\in\mathbb{R}^{N\times{}d}$ denote query, key, and value embeddings respectively. $W_q,W_k,W_v,W_o\in\mathbb{R}^{C\times{}C}$ are the projection matrices. To build up a Transformer block, an MLP block with two linear transformations and a GELU activation is usually adopted to provide nonlinearity.

With normalization layers and identity shortcuts, the $l$-th Transformer block is formulated as
\begin{align}
    z_l'&=\text{MHSA}\left(\text{LN}(z_{l-1})\right)+z_{l-1}, \label{eq:att_layer} \\
    z_l&=\text{MLP}\left(\text{LN}(z_l')\right)+z_l', \label{eq:mlp_layer}
\end{align}
where LN is Layer Normalization \cite{LN}. 
%erran: And the MHSA computational complexity is $\Omega(\text{MHSA})=2N^2C+4NC^2$, as $\mathcal{O}(N^2)$ w.r.t. sequence length $N$.
% The MHSA computational complexity is $\Omega(\text{MHSA})=2N^2C+4NC^2$.

% The Vision Transformer (ViT) \cite{ViT16x16} stacks multiple Transformer blocks to process the tokens sequentially, achieving a convolution-free model for image classification. However, the single-scale architecture in ViT and the quadratic computational complexity of MHSA pose limitations on the tasks requiring multiple resolution feature maps, such as object detection and instance segmentation. Pyramid Vision Transformer (PVT) \cite{PVT} and Swin Transformer \cite{Swin} introduce two different methods to tackle these difficulties. PVT first reshapes the $k_l$ and $v_l$ to $H\times{}W\times{}C$ and then downsamples them to a lower resolution $\frac{H}{r}\times{}\frac{W}{r}\times{}C$ by a factor $r$, resulting in a smaller computational complexity $O(\frac{N^2}{r^2})$. Swin Transformer proposes a window-based attention technique to perform attention in the separate $w\times{}w$ windows of the image tokens, which reduces the computational complexity to $O(w^2N)$. Besides the local window-based attention, shifting window attention is also adopted by to increase the receptive field progressively.

\subsection{Deformable Attention}

% DCN incorporate difficulty
% Rationale
% Deformable mechanism is first proposed by DCN \cite{DCNv1} in CNN to strengthen its capability on modeling geometric transformations in a data-dependent way, which allows a target element of feature map (query) to attend to a larger area (keys) than its original convolution offsets.

%erran: Existing hierarchical Vision Transformers try to address the challenge of excessive attention. The downsample technique in PVT \cite{PVT} results in severe information loss, and the shift-window attention in Swin Transformer \cite{Swin} leads to a much slower growth of receptive field, which limits modeling large objects. 
Existing hierarchical Vision Transformers, notably PVT \cite{PVT} and Swin Transformer \cite{Swin} try to address the challenge of excessive attention. The downsampling technique of the former results in severe information loss, and the shift-window attention of the latter leads to a much slower growth of receptive fields, which limits the potential of modeling large objects. 
Thus a data-dependent sparse attention is required to flexibly model relevant features, leading to deformable mechanism firstly proposed in DCN \cite{DCNv1}. However, simply implementing DCN in Transformer models is a non-trivial problem.
In DCN, each element on the feature map learns its offsets individually, of which a $3\!\times{}\!3$ deformable convolution on an $H\!\times{}\!W\!\times{}\!C$ feature map has the space complexity of $9HWC$. If we directly apply the same mechanism in  the attention module, the space complexity will drastically rise to $N_\text{q}N_\text{k}C$, where $N_\text{q}, N_\text{k}$ are the number of queries and keys and usually have the same scale as the feature map size $HW$, bringing approximately a biquadratic complexity.
% xzf:
Although Deformable DETR \cite{DeformableDETR} has managed to reduce this overhead by setting a lower number of keys with $N_\text{k}\!=\!4$ at each scale and works well as a detection head, 
%erran: it is infeasible to attend to such few keys in a backbone network for the unacceptable loss
it is inferior to attend to such few keys in a backbone network because of the unacceptable loss
of information (see detailed comparison in Appendix). In the meantime, the observations in \cite{deepvit, gcnet} have revealed that different queries have similar attention maps in visual attention models. Therefore, we opt for a simpler solution with shared shifted keys and values for each query to achieve an efficient trade-off.
%Since a naive adaptation is infeasible,
% Therefore, we instead opt for simpler deformable attention with shared shifted keys and values for each query, based on the observation in \cite{deepvit, gcnet} that different queries have similar attention maps in visual attention models

% as different queries have similar attention maps in visual attention models \cite{deepvit, gcnet}.

Specifically, we propose deformable attention to model the relations among tokens effectively under the guidance of the important regions in the feature maps. These focused regions are determined by multiple groups of deformed sampling points which are learned from the queries by an offset network. 
% Inspired by DCN \cite{DCNv1}, 
We adopt bilinear interpolation to sample features from the feature maps, and then the sampled features are fed to the key and value projections to get the deformed keys and values. Finally, standard multi-head attention is applied to attend queries to the sampled keys and aggregate features from the deformed values. Additionally, the locations of deformed points provide a more powerful relative position bias to facilitate the learning of the deformable attention, which will be discussed in the following sections. 

\noindent
\textbf{Deformable attention module.}
As illustrated in Figure \ref{fig:fig2}(a), given the input feature map $x\in\mathbb{R}^{H\times{}W\times{}C}$, a uniform grid of points $p\in\mathbb{R}^{H_G\times{}W_G\times{}2}$ are generated as the references.
Specifically, the grid size is downsampled from the input feature map size by a factor $r$, $H_G=H/r,W_G=W/r$. The values of reference points are linearly spaced 2D coordinates $\{(0,0),\ldots,(H_G-1,W_G-1)\}$, and then we normalize them to the range $[-1,+1]$ according to the grid shape $H_G\times{}W_G$, in which $(-1,-1)$ indicates the top-left corner and $(+1,+1)$ indicates the bottom-right corner. To obtain the offset for each reference point, the feature maps are projected linearly to the query tokens $q\!=\!xW_q$,
%where $W_q\!\in\!\mathbb{R}^{C\times{}C}$ denotes the query projection. 
and then fed into a light weight sub-network $\theta_\text{offset}(\cdot)$ to generate the offsets $\Delta{}p=\theta_\text{offset}(q)$.
% , where $\Delta{}p$ have the same shape as the reference points $p$. 
To stabilize the training process, we scale the amplitude of $\Delta{}p$ by some predefined factor $s$ to prevent too large offset, \textit{i.e.,} $\Delta{}p\!\xleftarrow{}\!s\tanh{(\Delta{}p)}$. Then the features are sampled at the locations of deformed points as keys and values, followed by projection matrices:
\begin{align}
    q=&xW_q,\ \tilde{k}=\tilde{x}W_k,\ \tilde{v}=\tilde{x}W_v, \label{eq:proj_dmha} \\
    \text{with} \ &\Delta{}p=\theta_\text{offset}(q), \ \tilde{x}=\phi(x;p+\Delta{}p). \label{eq:sampling}
\end{align}
$\tilde{k}$ and $\tilde{v}$ represent the deformed key and value embeddings respectively.
Specifically, we set the sampling function $\phi(\cdot;\cdot)$ to a bilinear interpolation to make it differentiable:
\begin{equation}
    \phi\left(z;(p_x,p_y)\right)\!=\!\sum_{(r_x,r_y)}\!g(p_x,r_x)g(p_y,r_y)z[r_y,r_x,:], \label{eq:bilinear}
\end{equation}
% where $\phi(z;p)$ is a sample operation, which gives the sampled feature value on $z$ according to the specified locations $p$. 
% Since the locations of deformed points are fractional, 
%  Let $(p_x,p_y)$ be the location to sample and $z$ is the feature map, then
where $g(a,b)=\max(0,1-|a-b|)$ and $(r_x,r_y)$ indexes all the locations on $z\!\in\!\mathbb{R}^{H\times{}W\times{}C}$. As $g$ would be non-zero only on the 4 integral points closest to $(p_x,p_y)$, it simplifies Eq.\eqref{eq:bilinear} to a weighted average on 4 locations. 
% After $x'$ is sampled, the keys and values are obtained as $k=x'W_k, v=x'W_v,$ where $W_k, W_v\in\mathbb{R}^{C\times{}C}$ are the key and value projections. %which are usually implemented as $1\times{}1$ convolutions in practice as well as $W_q$ and $W_o$. 
Similar to existing approaches, we perform multi-head attention on $q,k,v$ and adopt relative position offsets $R$. The output of an attention head is formulated as:
\begin{equation}
    z^{(m)}=\sigma\left(q^{(m)}\tilde{k}^{(m)\top}/\sqrt{d}+\phi(\hat{B};R)\right)\tilde{v}^{(m)},
\end{equation}
where $\phi(\hat{B};R)\in\mathbb{R}^{HW\times{}H_GW_G}$ correspond to the position embedding following previous work \cite{Swin} while with several adaptations. Details will be explained later in this section. Features of each head are concatenated together and projected through $W_o$ to get the final output $z$ as Eq.\eqref{eq:MHSA}.
%erran:
% Projections $W_k$, $W_v$, $W_q$ and $W_o$ are usually implemented as $1\times{}1$ convolutions in practice.
\begin{figure*}
    \centering
    \includegraphics[width=\linewidth]{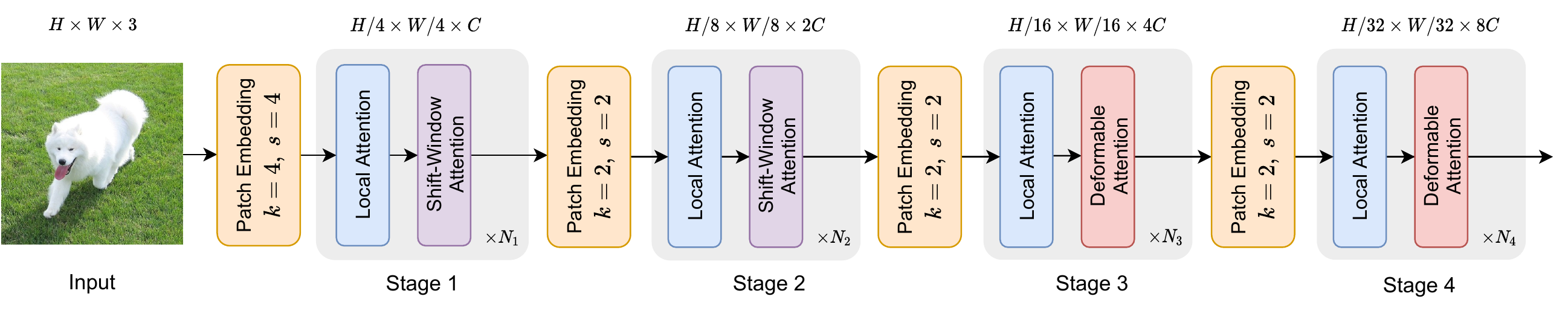}
    \caption{An illustration of \ourmethod{} architecture. $N_1$ to $N_4$ are the numbers of stacked successive local attention and shift-window / deformable attention blocks. $k$ and $s$ denote the kernel size and stride of the convolution layer in patch embeddings. }
    \label{fig:fig3}
\end{figure*}

\noindent
\textbf{Offset generation.} 
% In our deformable attention module, each query shares the same sets of deformable keys and values, which are sampled all over the feature maps. To make these points represent the important parts of the features, each offset should be aware of a large region surrounding its reference point and different offsets should be different according to the locations of their references. A three layers convolution module is adopted to meet these requirements, as illustrated in Figure \ref{fig:fig2}(b). A $k\times{}k$ depth-wise convolution first downsamples the query tokens to increase the receptive field of each offset feature by $k$, and then an activation unit is applied followed by a $1\times{}1$ convolution without bias to get the 2D offsets. 
As we have stated, a sub-network is adopted for offset generation which consumes the query features and outputs the offset values for reference points respectively. Considering that each reference point covers a local $s\times s$ region ($s$ is the largest value for offset), the generation network should also have the perception of the local features to learn reasonable offsets. Therefore, we implement the sub-network as two convolution modules with a nonlinear activation, as depicted in Figure \ref{fig:fig2}(b). The input features are first passed through a $5\!\times \!5$ depthwise convolution to capture local features. Then, GELU activation and a $1\!\times \!1$ convolution is adopted to get the 2D offsets. It is also worth noticing that the bias in $1\times{}1$ convolution is dropped to alleviate the compulsive shift for all locations. 

\noindent
\textbf{Offset groups.} 
% To promote the diversity of the deformed points, there are $G$ groups of learnable offsets for each deformable attention module. Thus the query should be chunked in the feature dimension $C$ to $C_G=C/G$, which comes to $C_G$ groups of offsets. In practice, offset groups $G$ is usually larger than attention heads $M$, or $C_G$ is several times of $d$, thus multiple attention heads are assigned to one group of deformed keys and values.
To promote the diversity of the deformed points, we follow a similar paradigm in MHSA, and split the feature channel into $G$ groups. Features from each group use the shared sub-network to generate the corresponding offsets respectively. In practice, the head number $M$ for the attention module is %set as in 
%set to be times of the offset groups $G$, 
set to be multiple times of the size of offset groups $G$,
ensuring that multiple attention heads are assigned to one group of deformed keys and values.

\noindent
\textbf{Deformable relative position bias.} Relative position bias encodes the relative positions between every pair of query and key, which augments the vanilla attention with spatial information. Considering a feature map with shape $H\!\times{}\!W$, its relative coordinate displacements lie in the range of $[-H,H]$ and $[-W,W]$ at two dimensions respectively. In Swin Transformer \cite{Swin}, a relative position bias table $\hat{B}\in\mathbb{R}^{(2H-1)\times{}(2W-1)}$ is constructed to obtain the relative position bias $B$ by indexing the table with the relative displacements in two directions. Since our deformable attention has continuous positions of keys, we compute the relative displacements in the normalized range $[-1,+1]$, and then interpolate $\phi(\hat{B};R)$ in the parameterized bias table $\hat{B}\in\mathbb{R}^{(2H-1)\times{}(2W-1)}$ by the continuous relative displacements in order to cover all possible offset values.

\noindent
\textbf{Computational complexity.} Deformable multi-head attention (DMHA) has a similar computation cost as the counterpart in PVT or Swin Transformer. The only additional overhead comes from the sub-network that is used to generate offsets.
The complexity of the whole module can be summarized as:
% Deformable multi-head attention (DMHA) includes 4 linear projections, 2 matrix multiplications, 2 bilinear interpolations and a convolution module for offset generation. Therefore, the complexity of an $M$-head DMHA is 
% \begin{equation}
%     \begin{aligned}
%         \Omega(\text{DMHA})&=2HWN_\text{s}C+2HWC^2+2N_\text{s}C^2\\
%         &+4N_\text{s}C+4MHWN_\text{s}+(k^2+2)N_\text{s}C,
%     \end{aligned}
%     \label{eq:dmha_comp}
% \end{equation}
\begin{equation}
    \Omega(\text{DMHA})\!=\!\underbrace{2HWN_\text{s}C\!\!+\!\!2HWC^2\!\!+\!\!2N_\text{s}C^2}_{\text{vanilla self-attention module}}\!+\!\underbrace{(k^2\!+\!2)N_\text{s}C}_{\text{\textcolor{blue}{offset network}}},
\end{equation}
% \begin{equation}
%     \begin{aligned}
%         \Omega(\text{DMHA})&=2HWN_\text{s}C+2HWC^2+2N_\text{s}C^2\\
%         &+(k^2+2)N_\text{s}C,
%     \end{aligned}
%     \label{eq:dmha_comp}
% \end{equation}
where $N_\text{s}\!=\!H_GW_G\!=\!HW/r^2$ is the number of sampled points. It can be immediately seen that the computational cost of the offset network has linear complexity \textit{w.r.t.} the channel size, which is comparably minor to the cost for attention computation. Typically, consider the third stage of a Swin-T \cite{Swin} model for image classification where $H\!=\!W\!=\!14$, $N_s\!=\!49$, $C\!=\!384$, the computational cost for the attention module in a single block is 79.63M FLOPs. If equipped with our deformable module (with $k\!=\!5$), the additional overhead is 5.08M Flops, which is only $6.0\%$ of the whole module. Additionally, by choosing a large downsample factor $r$, the complexity will be further reduced, which makes it friendly to the tasks with much higher resolution inputs such as object detection and instance segmentation.

% If we choose a large downsample factor $r$ or set $N_\text{s}\ll{}HW$ to a fixed value, the complexity will be reduced significantly, which makes it friendly to the tasks with much higher resolution maps such as object detection and instance segmentation.

\subsection{Model Architectures}
We replace the vanilla MHSA with our deformable attention in the Transformer (Eq.\eqref{eq:att_layer}), and combine it with an MLP (Eq.\eqref{eq:mlp_layer}) to build a deformable vision transformer block. In terms of the network architecture, our model, \textbf{Deformable Attention Transformer}, shares a similar pyramid structure with \cite{PVT,Swin,dpt,botnet}, which is 
%erran: accommodable at various vision tasks which need high resolution feature maps
broadly applicable to various vision tasks requiring multiscale feature maps. As illustrated in Figure \ref{fig:fig3}, an input image with shape $H\times{}W\times{}3$ is firstly embedded by a 4$\times{}$4 non-overlapped convolution with stride 4, followed by a normalization layer to get the $\frac{H}{4}\times{}\frac{W}{4}\times{}C$ patch embeddings. Aiming to build a hierarchical feature pyramid, the backbone includes 4 stages with a progressively increasing stride. Between two consecutive stages, there is a non-overlapped 2$\times{}$2 convolution with stride 2 to downsample the feature map to halve the spatial size and double the feature dimensions. In classification task, we first normalize the feature maps output from the last stage and then adopt a linear classifier with pooled features to predict the logits. In object detection, instance segmentation and semantic segmentation tasks, DAT plays the role of a backbone in an integrated vision model to extract multiscale features. We add a normalization layer to the features from each stage before feeding them into the following modules such as FPN \cite{fpn} in object detection or decoders in semantic segmentation.

% Swin justification
% local features at early stages
We introduce successive local attention and deformable attention blocks in the third and the fourth stage of DAT. The feature maps are firstly processed by a window-based local attention to aggregate information locally, and then passed through the deformable attention block to model the global relations between the local augmented tokens. This alternate design of attention blocks with local and global receptive fields helps the model learn strong representations, sharing a similar pattern in GLiT \cite{glit}, TNT \cite{tnt} and Pointformer \cite{pointformer}. 
%erran: Since the features in the first two stages are too local to guide learning offsets, deformable attentions in the early stages would trap the optimization and dim the model performance. 
Since the first two stages mainly learn local features, deformable attention in these early stages is less preferred.
In addition, the keys and values in the first two stages have a rather large spatial size, which greatly increase the computational overhead in the dot products and bilinear interpolations in deformable attention. Therefore, to achieve a trade-off between model capacity and computational burden, we only place deformable attention in the third and the fourth stage and adopt the shift-window attention in Swin Transformer \cite{Swin} to have a better representation in the early stages. We build three variants of DAT in different parameters and FLOPs for a fair comparison with other Vision Transformer models. We change the model size by stacking more blocks in the third stage and increasing the hidden dimensions. The detailed architectures are reported in Table \ref{tab:arch}.
%erran: It is also worth noticing that the first two stages of DAT can adapt to different self-attention mechanisms in Vision Transformer models, 
% different self-attention mechanisms in Vision Transformer models, 
Note that there are other design choices for the first two stages of DAT, 
e.g., the SRA module in PVT. We show the comparison results in Table \ref{tab:abl_ds}.

\begin{table}[t]
\newcommand{\tabincell}[2]{\begin{tabular}{@{}#1@{}}#2\end{tabular}}
\begin{center}
\setlength{\tabcolsep}{0.01mm}{
\renewcommand\arraystretch{1.0}
\begin{tabular}{c|c|c|c}
\thickhline
\multicolumn{4}{c}{\textbf{\ourmethod{} Architectures}} \\
 & \ourmethod-T & \ourmethod-S & \ourmethod-B \\
\hline
\multirow{3}{*}{
$\begin{matrix}\text{Stage 1}\\(\text{56}\!\times{}\!\text{56})\end{matrix}$
} 
& $N_1\!=\!1,C\!=\!96$ & $N_1\!=\!1,C\!=\!96$ & $N_1\!=\!1,C\!=\!128$ \\
& window size: 7 & window size: 7 & window size: 7 \\
& heads: 3 & heads: 3 & heads: 4 \\
\hline
\multirow{3}{*}{
$\begin{matrix}\text{Stage 2}\\(\text{28}\!\times{}\!\text{28})\end{matrix}$
} 
& $N_2\!=\!1,C\!=\!192$ & $N_2\!=\!1,C\!=\!192$ & $N_2\!=\!1,C\!=\!256$ \\
& window size: 7 & window size: 7 & window size: 7 \\
& heads: 6 & heads: 6 & heads: 8\\
\hline
\multirow{4}{*}{
$\begin{matrix}\text{Stage 3}\\(\text{14}\!\times{}\!\text{14})\end{matrix}$
} 
& $N_3\!=\!3,C\!=\!384$ & $N_3\!=\!9,C\!=\!384$ & $N_3\!=\!9,C\!=\!512$ \\
& window size: 7 & window size: 7 & window size: 7 \\
& heads: 12 & heads: 12 & heads: 16\\
& groups: 3 & groups: 3 & groups: 4 \\
\hline
\multirow{4}{*}{
$\begin{matrix}\text{Stage 4}\\(\text{7}\!\times{}\!\text{7})\end{matrix}$
} 
& $N_4\!=\!1,C\!=\!768$ & $N_4\!=\!1,C\!=\!768$ & $N_4\!=\!1,C\!=\!1024$ \\
& window size: 7 & window size: 7 & window size: 7 \\
& heads: 24 & heads: 24 & heads: 32 \\
& groups: 6 & groups: 6 & groups: 8 \\
\thickhline
\end{tabular}}
\end{center}
\vskip -0.1in
\caption{Model architecture specifications. $\boldsymbol{N_i}$: Number of block at stage i. $\boldsymbol{C}$: Channel dimension. \textbf{window size}: Region size in local attention module. \textbf{heads}: Number of heads in DMHA. \textbf{groups}: Offset groups in DMHA.}
\label{tab:arch}
\vskip -0.2in
\end{table}
\section{Experiments}

We conduct experiments on several datasets to verify the effectiveness of our proposed \ourmethod. We show our results on ImageNet-1K \cite{in1k} classification, COCO \cite{coco} object detection and ADE20K \cite{ade20k} semantic segmentation tasks. In addition, we provide ablation studies and 
%some visualizations to further discuss our method.
visualizations to further show the effectiveness of our method.

\subsection{ImageNet-1K Classification}

ImageNet-1K \cite{in1k} dataset has 1.28M images for training and 50K images for validation. We train three variants of \ourmethod{} on the training split and report the Top-1 accuracy on the validation split to compare with other Vision Transformer models.

We use AdamW \cite{adamw} optimizer to train our models for 300 epochs with a cosine learning rate decay. The basic learning rate for a batch size of 1024 is set to $1\times10^{-3}$, and then linearly scaled w.r.t. the batch size. To stabilize training procedures, we schedule a linear warm-up of learning rate from $1\times10^{-6}$ to the basic learning rate, and for a better convergence the cosine decay rule is applied to gradually decrease the learning rate to $1\times10^{-7}$ during training. We follow DeiT \cite{deit} to set the advanced data augmentation, including RandAugment \cite{randaug}, Mixup \cite{mixup} and CutMix \cite{cutmix} to avoid overfitting. In addition, stochastic depth \cite{sdepth} and weight decay of 0.05 are also applied, in which the stochastic depth degree is chosen 0.2, 0.3 and 0.5 for the tiny, small and base model, respectively. We do not adopt EMA \cite{ema}, random erasing \cite{randerase} and the vanilla drop out, which 
%erran: contribute nothing to 
does not improve
the training of Vision Transformers, as verified in \cite{deit,Swin}. In terms of larger resolution finetuning, we finetune our \ourmethod{}-B using AdamW optimizer with a cosine scheduled learning rate $4\times{}10^{-6}$ for 30 epochs. 
%erran; We hold the stochastic depth rate of 0.5 and lower the weight decay to $1\times{}10^{-8}$ to keep a strong regularization. 
We set the stochastic depth rate to 0.5 and lower the weight decay to $1\times{}10^{-8}$ to keep the regularization.

We report our results in Table \ref{tab:cls}, with 300 training epochs. Compared with other state-of-the-art Vision Transformers, our \ourmethod{}s achieve significant improvements on the Top-1 accuracy with similar computational complexities. Our method \ourmethod{} outperforms Swin Transformer \cite{Swin}, PVT \cite{PVT}, DPT \cite{dpt} and DeiT \cite{deit} in all three scales. Without inserting convolutions in Transformer blocks \cite{pvtv2, xcit, cmt}, or using overlapped convolutions in patch embeddings \cite{rgvit, cswin, hrformer}, \ourmethod{}s achieve gains of +0.7, +0.7 and +0.5 over Swin Transformer \cite{Swin} counterparts. When finetuning at $384\times{}384$ resolution, our model continues performing better than Swin Transformer by 0.3.
%, which shows the larger capacity of \ourmethod{}. 

\begin{table}[t]
\newcommand{\tabincell}[2]{\begin{tabular}{@{}#1@{}}#2\end{tabular}}
\begin{center}
\setlength{\tabcolsep}{1.0mm}{
\renewcommand\arraystretch{1.0}
\begin{tabular}{l|ccc|c}
\thickhline
\multicolumn{5}{c}{\textbf{ImageNet-1K Classification}} \\
Method & Resolution & FLOPs & \#Param & Top-1 Acc.\\
\hline
DeiT-S \cite{deit} & $\text{224}^\text{2}$ & 4.6G & 22M & 79.8 \\
PVT-S \cite{PVT} & $\text{224}^\text{2}$ & 3.8G & 25M & 79.8 \\
GLiT-S \cite{glit} & $\text{224}^\text{2}$ & 4.4G & 25M & 80.5 \\
DPT-S \cite{dpt} & $\text{224}^\text{2}$ & 4.0G & 26M & 81.0 \\
Swin-T \cite{Swin} &  $\text{224}^\text{2}$ & 4.5G & 29M & 81.3 \\
\rowcolor{mygray}
\textbf{\ourmethod-T} & $\text{224}^\text{2}$ & 4.6G & 29M & \textbf{82.0}\ $\textbf{\scriptsize{(+0.7)}}$ \\
\hline
PVT-M \cite{PVT} & $\text{224}^\text{2}$ & 6.9G & 46M & 81.2 \\
PVT-L \cite{PVT} & $\text{224}^\text{2}$ & 9.8G & 61M & 81.7 \\
DPT-M \cite{dpt} & $\text{224}^\text{2}$ & 6.9G & 46M & 81.9 \\
Swin-S \cite{Swin} &  $\text{224}^\text{2}$ & 8.8G & 50M & 83.0 \\
\rowcolor{mygray}
\textbf{\ourmethod-S} &  $\text{224}^\text{2}$ & 9.0G & 50M & \textbf{83.7}\ $\textbf{\scriptsize{(+0.7)}}$ \\
\hline
DeiT-B \cite{deit} & $\text{224}^\text{2}$ & 17.5G & 86M & 81.8 \\
GLiT-B \cite{glit} & $\text{224}^\text{2}$ & 17.0G & 96M & 82.3 \\
% DeiT-B \cite{deit} & $\text{384}^\text{2}$ & 55.4G & 86M & 83.1 \\
Swin-B \cite{Swin} & $\text{224}^\text{2}$ & 15.5G & 88M & 83.5 \\
\rowcolor{mygray}
\textbf{\ourmethod-B} & $\text{224}^\text{2}$ & 15.8G & 88M & \textbf{84.0}\ $\textbf{\scriptsize{(+0.5)}}$\\
\hline
DeiT-B \cite{deit} & $\text{384}^\text{2}$ & 55.4G & 86M & 83.1 \\
Swin-B \cite{Swin} & $\text{384}^\text{2}$ & 47.2G & 88M & 84.5 \\
\rowcolor{mygray}
\textbf{\ourmethod-B} & $\text{384}^\text{2}$ & 49.8G & 88M & \textbf{84.8}\ $\textbf{\scriptsize{(+0.3)}}$ \\
\thickhline
\end{tabular}}
\end{center}
\caption{
%Comparisons of FLOPS and parameters against accuracy on ImageNet-1K classification task.
Comparisons of \ourmethod{} with other vision transformer backbones on FLOPs, parameters, accuracy on the ImageNet-1K classification task.
}
\label{tab:cls}
\end{table}

\begin{table}[t]
\begin{center}
\setlength{\tabcolsep}{0.2mm}{
\renewcommand\arraystretch{1.2}
\begin{tabular}{l|c|c|c|ccc|ccc}
\thickhline
\multicolumn{10}{c}{\textbf{RetinaNet Object Detection on COCO}} \\
Method & FLOPs & \#Param & Sch. & AP & AP$_\text{50}$ & AP$_\text{75}$ & AP$_{s}$ & AP$_{m}$ & AP$_{l}$ \\
\hline
PVT-S & 286G & 34M & 1x & 40.4 & 61.3 & 43.0 & 25.0 & 42.9 & 55.7 \\
Swin-T & 248G & 38M & 1x & 41.7 & 63.1 & 44.3 & 27.0 & 45.3 & 54.7 \\
\rowcolor{mygray}
\textbf{\ourmethod-T} & 253G & 38M & 1x & 42.8 & 64.4 & 45.2 & 28.0 & 45.8 & 57.8 \\
\hline
PVT-S & 286G & 34M & 3x & 42.3 & 63.1 & 44.8 & 26.7 & 45.1 & 57.2 \\
Swin-T & 248G & 38M & 3x & 44.8 & 66.1 & 48.0 & 29.2 & 48.6 & 58.6 \\
\rowcolor{mygray}
\textbf{\ourmethod-T} & 253G & 38M & 3x & 45.6 & 67.2 & 48.5 & 31.3 & 49.1 & 60.8 \\
\hline
Swin-S & 339G & 60M  & 1x & 44.5 & 66.1 & 47.4 & 29.8 & 48.5 & 59.1 \\
\rowcolor{mygray}
\textbf{\ourmethod-S} & 359G & 60M & 1x & 45.7 & 67.7 & 48.5 & 30.5 & 49.3 & 61.3 \\
\hline
Swin-S & 339G & 60M  & 3x & 47.3 & 68.6 & 50.8 & 31.9 & 51.8 & 62.1 \\
\rowcolor{mygray}
\textbf{\ourmethod-S} & 359G & 60M & 3x & 47.9 & 69.6 & 51.2 & 32.3 & 51.8 & 63.4 \\
\thickhline
\end{tabular}}
\end{center}
\caption{Results on COCO object detection with RetinaNet \cite{rtn}. The table displays the number of parameters, computational cost (FLOPs), mAP at different mIoU thresholds and different object sizes. The FLOPs are computed over backbone, FPN and detection head with RGB input image at the resolution of 1280$\times$800. }
\label{tab:det1}
\end{table}

\begin{table*}[t]
\begin{center}
\setlength{\tabcolsep}{1mm}{
\renewcommand\arraystretch{1.2}
\begin{tabular}{l|c|c|c|ccc|ccc|ccc|ccc}
\thickhline
\multicolumn{16}{c}{\textbf{(a) Mask R-CNN Object Detection \& Instance Segmentation on COCO}} \\
Method & FLOPs & \#Param & Schedule & AP$^b$ & AP$^b_\text{50}$ & AP$^b_\text{75}$ & AP$^b_s$ & AP$^b_m$ & AP$^b_l$ & AP$^m$ & AP$^m_\text{50}$ & AP$^m_\text{75}$ & AP$^m_s$ & AP$^m_m$ & AP$^m_l$ \\
\hline
Swin-T & 267G & 48M & 1x & 43.7 & 66.6 & 47.7 & 28.5 & 47.0 & 57.3 & 39.8 & 63.3 & 42.7 & 24.2 & 43.1 & 54.6 \\
\rowcolor{mygray}
\textbf{\ourmethod-T} & 272G & 48M & 1x & 44.4 & 67.6 & 48.5 & 28.3 & 47.5 & 58.5 & 40.4 & 64.2 & 43.1 & 23.9 & 43.8 & 55.5 \\ 
\hline
Swin-T & 267G & 48M & 3x & 46.0 & 68.1 & 50.3 & 31.2 & 49.2 & 60.1 & 41.6 & 65.1 & 44.9 & 25.9 & 45.1 & 56.9 \\
\rowcolor{mygray}
\textbf{\ourmethod-T} & 272G & 48M & 3x & 47.1 & 69.2 & 51.6 & 32.0 & 50.3 & 61.0 & 42.4 & 66.1 & 45.5 & 27.2 & 45.8 & 57.1 \\
\hline
Swin-S & 359G & 69M & 1x & 45.7 & 67.9 & 50.4 & 29.5 & 48.9 & 60.0 & 41.1 & 64.9 & 44.2 & 25.1 & 44.3 & 56.6 \\
\rowcolor{mygray}
\textbf{\ourmethod-S} & 378G & 69M & 1x & 47.1 & 69.9 & 51.5 & 30.5 & 50.1 & 62.1 & 42.5 & 66.7 & 45.4 & 25.5 & 45.8 & 58.5 \\
\hline
Swin-S & 359G & 69M & 3x & 48.5 & 70.2 & 53.5 & 33.4 & 52.1 & 63.3 & 43.3 & 67.3 & 46.6 & 28.1 & 46.7 & 58.6 \\
\rowcolor{mygray}
\textbf{\ourmethod-S} & 378G & 69M & 3x & 49.0 & 70.9 & 53.8 & 32.7 & 52.6 & 64.0 & 44.0 & 68.0 & 47.5 & 27.8 & 47.7 & 59.5 \\
\thickhline
\multicolumn{16}{c}{\textbf{(b) Cascade Mask R-CNN Object Detection \& Instance Segmentation on COCO}} \\
Method & FLOPs & \#Param & Schedule & AP$^b$ & AP$^b_\text{50}$ & AP$^b_\text{75}$ & AP$^b_s$ & AP$^b_m$ & AP$^b_l$ & AP$^m$ & AP$^m_\text{50}$ & AP$^m_\text{75}$ & AP$^m_s$ & AP$^m_m$ & AP$^m_l$ \\
\hline
Swin-T & 745G & 86M & 1x & 48.1 & 67.1 & 52.2 & 30.4 & 51.5 & 63.1 & 41.7 & 64.4 & 45.0 & 24.0 & 45.2 & 56.9 \\
\rowcolor{mygray}
\textbf{\ourmethod-T} & 750G & 86M & 1x & 49.1 & 68.2 & 52.9 & 31.2 & 52.4 & 65.1 & 42.5 & 65.4 & 45.8 & 25.2 & 45.9 & 58.6 \\ 
\hline
Swin-T & 745G & 86M & 3x & 50.4 & 69.2 & 54.7 & 33.8 & 54.1 & 65.2 & 43.7 & 66.6 & 47.3 & 27.3 & 47.5 & 59.0 \\
\rowcolor{mygray}
% \textbf{\ourmethod-T} & 755G & 86M & 3x & 51.4 & 70.3 & 55.7 & 36.3 & 54.8 & 66.5 & 44.7 & 67.8 & 48.7 & 29.7 & 48.2 & 60.0 \\
\textbf{\ourmethod-T} & 750G & 86M & 3x & 51.3 & 70.1 & 55.8 & 34.1 & 54.6 & 66.9 & 44.5 & 67.5 & 48.1 & 27.9 & 47.9 & 60.3 \\
\hline
Swin-S & 838G & 107M & 3x & 51.9 & 70.7 & 56.3 & 35.2 & 55.7 & 67.7 & 45.0 & 68.2 & 48.8 & 28.8 & 48.7 & 60.6 \\
\rowcolor{mygray}
\textbf{\ourmethod-S} & 857G & 107M & 3x & 52.7 & 71.7 & 57.2 & 37.3 & 56.3 & 68.0 & 45.5 & 69.1 & 49.3 & 30.2 & 49.2 & 60.9 \\
\hline
Swin-B & 982G & 145M & 3x & 51.9 & 70.5 & 56.4 & 35.4 & 55.2 & 67.4 & 45.0 & 68.1 & 48.9 & 28.9 & 48.3 & 60.4 \\
\rowcolor{mygray}
\textbf{\ourmethod-B} & 1003G & 145M & 3x & 53.0 & 71.9 & 57.6 & 36.0 & 56.8 & 69.1 & 45.8 & 69.3 & 49.5 & 29.2 & 49.5 & 61.9 \\
\thickhline
\end{tabular}}
\end{center}
\caption{Results on COCO object detection and instance segmentation. The table displays the number of parameters, computational cost (FLOPs), mAP at different IoU thresholds and mAP for objects in different sizes. The FLOPs are computed over backbone, FPN and detection head with RGB input image at the resolution of 1280$\times$800. }
\label{tab:det2}
\vskip -0.1in
\end{table*}

\subsection{COCO Object Detection}

COCO \cite{coco} object detection and instance segmentation dataset has 118K training images and 5K validation images. We use our \ourmethod{} as the backbone in RetinaNet \cite{rtn}, Mask R-CNN \cite{mrcn} and Cascade Mask R-CNN \cite{cmrcn} frameworks to evaluate the effectiveness of our method. We pretrain our models on ImageNet-1K dataset for 300 epochs and follow the similar training strategies in Swin Transformer \cite{Swin} to compare our methods fairly.
% , except for larger stochastic detph rates for \ourmethod{}-S (0.3) and \ourmethod{}-B (0.5).

We report our \ourmethod{} on RetinaNet model in 1x and 3x training schedules. As shown in Table \ref{tab:det1}, \ourmethod{} outperforms Swin Transformer by 1.1 and 1.2 mAP among tiny and small models. When implemented in two-stage detectors, e.g., Mask R-CNN and Cascade Mask R-CNN, our model achieves consistent improvements over Swin Transformer models in different sizes, as shown in Table \ref{tab:det2}. We can see that \ourmethod{} achieves most improvements on large objects (up to +2.1) due to the flexibility in modeling long-range dependencies. The gaps for small objects detection and instance segmentation are also pronounced (up to +2.1), which shows that \ourmethod{}s also have the capacity of modeling relations in the local region.

% \textbf{NOT completed! MRCN / CMRCN comparison !}

\subsection{ADE20K Semantic Segmentation}

ADE20K \cite{ade20k} is a popular dataset for semantic segmentation with 20K training images and 2K validation images. We employ our \ourmethod{} on two widely adopted segmentation models, SemanticFPN \cite{semfpn} and UperNet \cite{upernet}. To make a fair comparison to PVT \cite{PVT} and Swin Transformer \cite{Swin}, we follow the learning rate schedules and training epochs, except for the degree of stochastic depth, which is a key hyper-parameter affecting the final performance. We set it for 0.3, 0.3 and 0.5 for tiny, small and base variants of our \ourmethod{} respectively for both two models. With the pretraining models on ImageNet-1K, we train SemanticFPN for 40k steps and UperNet for 160k steps. In Table \ref{tab:seg}, we report the results on the validation set with the highest mIoU score of all methods. In comparison with PVT \cite{PVT}, our tiny model outperforms PVT-S by +0.5 mIoU even with less FLOPs and achieves a sharp boost with +3.1 and +2.5 in mIoU with a slightly larger model size. Our \ourmethod{} has a significant improvement over the Swin Transformer at each of three model scales, with +1.0, +0.7 and +1.2 in mIoU respectively, showing our method's 
%erran: reviewers may not like this: superiority.
effectiveness.

\subsection{Ablation Study}

In this section, we ablate the key components in our \ourmethod{} to verify the effectiveness of these designs. We report the results on ImageNet-1K classification based on DAT-T.

\noindent
\textbf{Geometric information exploitation. } 
%The primary component of \ourmethod{} is the deformable attention to learn geometric transformation. 
We first evaluate the effectiveness of our proposed deformable offsets and deformable relative position embeddings, as shown in Table \ref{tab:abl_geo}. 
%We can see that performing full attention on the last two stages only gives +0.1 gain over Swin-T model. However 
Either adopting offsets in feature sampling or using deformable relative position embedding provides +0.3 improvement. We also try other types of position embeddings, including a fixed learnable position bias and a depth-wise convolution in \cite{cswin}. But none of them is effective with only +0.1 gain over that without position embedding, which shows our deformable relative position bias is more compatible with deformable attention. There is also an observation from rows 6 and 7 in Table \ref{tab:abl_geo} that our model can adapt to different attention modules at the first two stages and achieve competitive results. %For example, 
Our model with SRA \cite{PVT} at the first two stages outperforms PVT-M by 0.5 with 65$\%$ FLOPs.

\noindent
\textbf{Deformable attention at different stages. } 
We replace the shift-window attention of Swin Transfomer \cite{Swin} with our deformable attention at different stages. As shown in Table \ref{tab:abl_ds}, only replacing the attention in the last stage improves by 0.1 and replacing the last two stages leads to a performance gain of 0.7 (achieving an overall accuracy of 82.0). However, replacing with more deformable attention at the early stages slightly decreases the accuracy.

\begin{table}[t]
\newcommand{\tabincell}[2]{\begin{tabular}{@{}#1@{}}#2\end{tabular}}
\begin{center}
\setlength{\tabcolsep}{0.6mm}{
\renewcommand\arraystretch{1.2}
\begin{tabular}{c|c|cc|cc|c}
\thickhline
\multicolumn{7}{c}{\textbf{Semantic Segmentation on ADE20K}} \\
Backbone & Method & FLOPs & \#Params & mIoU & mAcc & mIoU$^\dagger$\\
\hline
PVT-S & S-FPN & 225G & 28M & 41.95 & 53.02 & 41.95 \\
\rowcolor{mygray}
\textbf{\ourmethod-T} & S-FPN & 198G & 32M & \textbf{42.56} & 54.72 & 44.22 \\
\hline
PVT-M & S-FPN & 315G & 48M & 42.91 & 53.80 & 43.34 \\
\rowcolor{mygray}
\textbf{\ourmethod-S} & S-FPN & 320G & 53M & \textbf{46.08} & 58.17 & 48.46 \\
\hline
PVT-L & S-FPN & 420G & 65M & 43.49 & 54.62 & 43.92 \\
\rowcolor{mygray}
\textbf{\ourmethod-B} & S-FPN & 481G & 92M & \textbf{47.02} & 59.47 & 49.01 \\
\hline
Swin-T & UperNet & 945G & 60M & 44.51 & 55.61 & 45.81 \\
\rowcolor{mygray}
\textbf{\ourmethod-T} & UperNet & 957G & 60M & \textbf{45.54} & 57.95 & 46.44 \\
\hline
Swin-S & UperNet & 1038G & 81M & 47.64 & 58.78 & 49.47 \\
 \rowcolor{mygray}
\textbf{\ourmethod-S} & UperNet & 1079G & 81M & \textbf{48.31}& 60.44 & 49.84  \\
\hline
Swin-B & UperNet & 1188G & 121M	& 48.13 & 59.13 & 49.72 \\
 \rowcolor{mygray}
\textbf{\ourmethod-B} & UperNet & 1212G & 121M & \textbf{49.38} &  61.82 & 50.55 \\
\thickhline
\end{tabular}}
\end{center}
\vskip -0.1in
\caption{Results of semantic segmentation. The FLOPs are computed over encoders and decoders with RGB input image at the resolution of 512$\times$2048. $\dagger$ denotes the metrics are reported under a multi-scale test setting with flip augmentation. S-FPN is short for SemanticFPN \cite{semfpn} model. The results of PVT and Swin Transformer are copied from their Github repositories, which are higher than the versions in their original papers.}
\label{tab:seg}
\end{table}

\begin{table}[t]
\newcommand{\tabincell}[2]{\begin{tabular}{@{}#1@{}}#2\end{tabular}}
\begin{center}
\setlength{\tabcolsep}{0.9mm}{
\renewcommand\arraystretch{1.2}
\begin{tabular}{ccc|cc|cc}
\thickhline
Attn. & Offsets & Pos. Emebd & FLOPs & \#Param & Acc. & Diff. \\
\hline
S & \ding{55} & \ding{55} & 4.57G & 28.29M & 81.4 & -0.6 \\
S & \ding{55} & \textbf{Relative} & 4.57G & 28.32M & 81.7 & -0.3 \\
\hline
S & \ding{51} & \ding{55} & 4.58G & 28.29M & 81.7 & -0.3 \\
S & \ding{51} & Fixed & 4.58G & 29.73M & 81.8 & -0.2 \\
S & \ding{51} & DWConv & 4.59G & 28.31M & 81.8 & -0.2 \\
\hline
% \multicolumn{3}{c|}{PVT-S \cite{PVT}} & 3.8G & 25M & 79.8 & -2.2 \\
P & \ding{51} & \textbf{Relative} & 4.48G & 30.68M & 81.7 & -0.3 \\
S & \ding{51} & \textbf{Relative} & 4.59G & 28.32M & 82.0 & \ourmethod{} \\
\thickhline
\end{tabular}}
\end{center}
\vskip -0.1in
\caption{Ablation study on different ways to exploiting geometric information. \textbf{P} represents the first two stages use SRA attention in \cite{PVT}, and \textbf{S} represents shift-window attention in \cite{Swin}. \ding{51} in offsets means performing spatial sampling in deformable attention module while \ding{55} means not.}
\label{tab:abl_geo}
\vskip -0.1in
\end{table}

\begin{table}[t]
\newcommand{\tabincell}[2]{\begin{tabular}{@{}#1@{}}#2\end{tabular}}
\begin{center}
\setlength{\tabcolsep}{0.9mm}{
\renewcommand\arraystretch{1.2}
\begin{tabular}{cccc|cc|c}
\thickhline
\multicolumn{4}{c|}{Stages w/ Deformable Attention} & \multirow{2}{*}{FLOPs} & \multirow{2}{*}{\#Param} & \multirow{2}{*}{Acc.} \\
Stage 1 & Stage 2 & Stage 3 & Stage 4 & & & \\
\hline
\ding{51} & \ding{51} & \ding{51} & \ding{51} & 4.64G & 28.39M & 81.7 \\
 & \ding{51} & \ding{51} & \ding{51} & 4.60G & 28.34M & 81.9 \\
 & & \ding{51} & \ding{51} & 4.59G & 28.32M & 82.0 \\
 & & & \ding{51} & 4.51G & 28.29M & 81.4 \\
\hline
\multicolumn{4}{c|}{Swin-T \cite{Swin}} & 4.51G & 28.29M & 81.3\\
\thickhline
\end{tabular}}
\end{center}
\vskip -0.1in
\caption{Ablation study on applying deformable attention on different stages. \ding{51} means this stage is made up of successive local attention and deformable attention Transformer blocks. Note that our model takes the relative position indices of all local and shift-window attention and the reference grid points of all deformable attention into parameter counting, which may lead to a higher number of parameters.}
\label{tab:abl_ds}
\end{table}

\begin{figure}
    \centering
    \includegraphics[width=\linewidth]{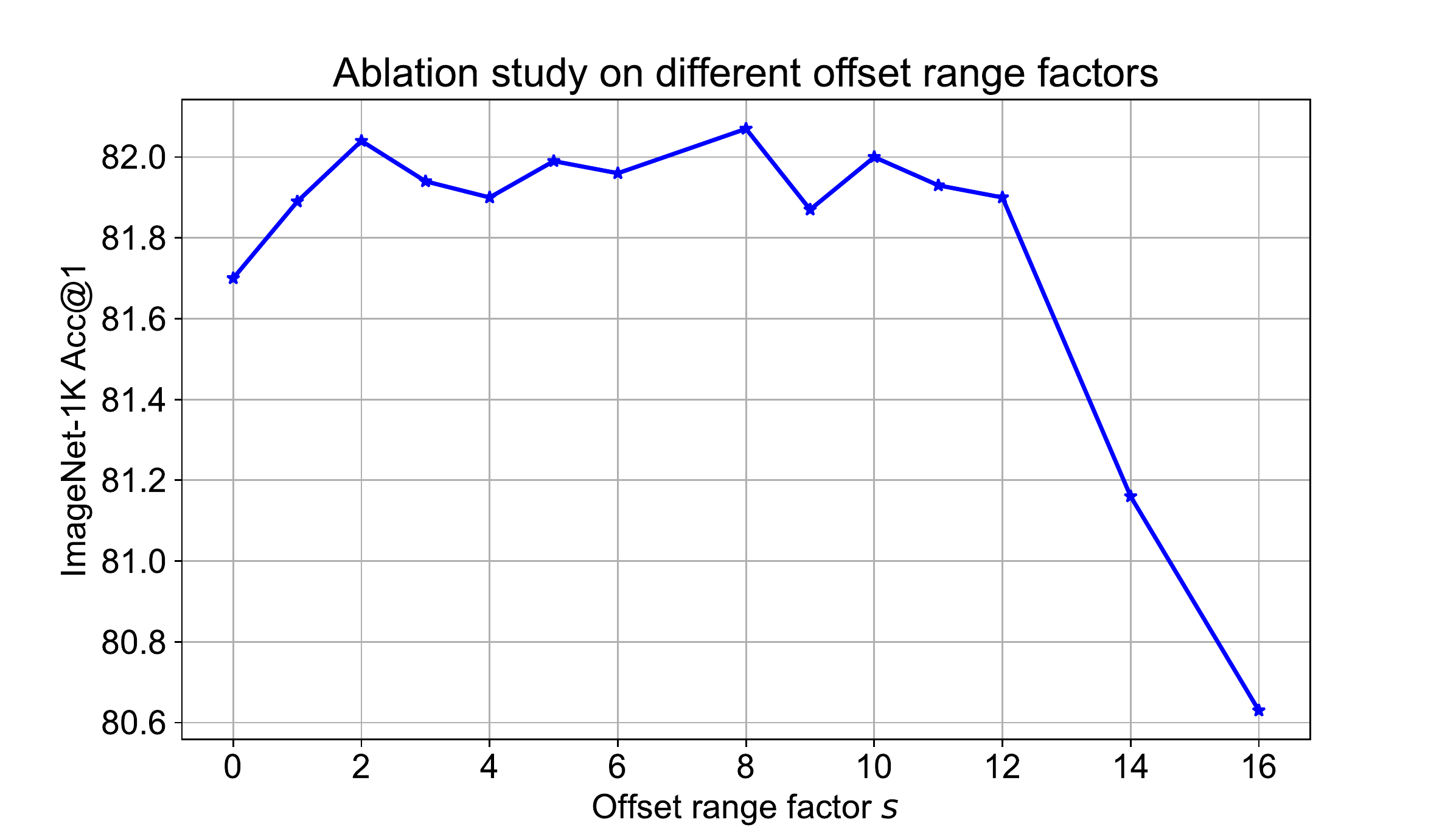}
    \caption{Ablation study on different offset range factor $s$. Accuracies of DAT-T on ImageNet show a wide range of $s$, implying the robustness of this hyper-parameter.}
    \label{fig:5}
    \vskip -0.1in
\end{figure}

% \begin{table}[t]
% \newcommand{\tabincell}[2]{\begin{tabular}{@{}#1@{}}#2\end{tabular}}
% \begin{center}
% \setlength{\tabcolsep}{0.9mm}{
% \renewcommand\arraystretch{1.2}
% \begin{tabular}{cc|cc|cc}
% \thickhline
% $G$ at Stage 3 & $G$ at Stage 4 & FLOPs & \#Params & Acc. & Diff. \\
% \hline
% 1 & 1 & 4.59G & 28.56M & 81.8 & -0.2 \\
% 6 & 12 & 4.59G & 28.52M & 81.9 & -0.1 \\
% 12 & 24 & 4.59G & 28.50M & 81.8 & -0.2 \\
% \hline
% 3 & 6 & 4.59G & 28.52M & 82.0 & \ourmethod{} \\
% \thickhline
% \end{tabular}}
% \end{center}
% \caption{Ablation study on number of groups. \ding{51} means this stage is made up of successive local attention and deformable attention Transformer blocks. Note that our model takes the relative position indices of all local and shift-window attention and the reference grid points of all deformable attention into parameter counting, which may lead to a higher number of parameters.}
% \label{tab:abl_grp}
% \end{table}

\noindent
\textbf{Ablation on different $s$.} We go on the further study of the impact of different maximum offsets, \textit{i.e.}, the offset range scale factor $s$ in the paper. We conduct an ablation experiment of $s$ ranging from 0 to 16 where 14 corresponds to the largest reasonable offset given the size of the feature map ($14\times{}14$ at stage 3). As shown in Figure~\ref{fig:5}, the wide selection range of $s$ shows the robustness of DAT to this hyper-parameter. Practically, we choose a small $s\!=\!2$ for all models in the paper without additional tuning.

%\subsection{Discussion and Visualization}
\subsection{Visualization}

\begin{figure}
	\centering
	\subfloat{\label{fig:fig6_1}\includegraphics[width=0.32\linewidth]{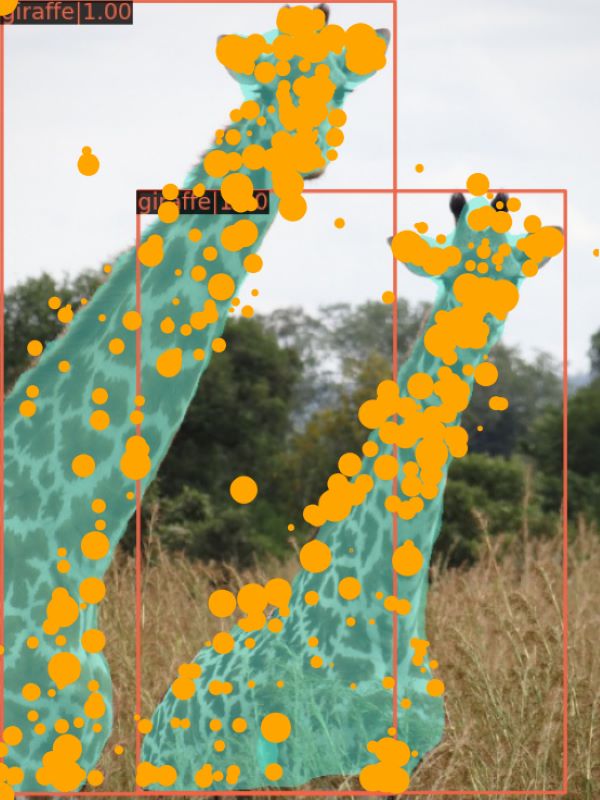}}
	\hspace{0.1pt}
    \subfloat{\label{fig:fig6_2}\includegraphics[width=0.64\linewidth]{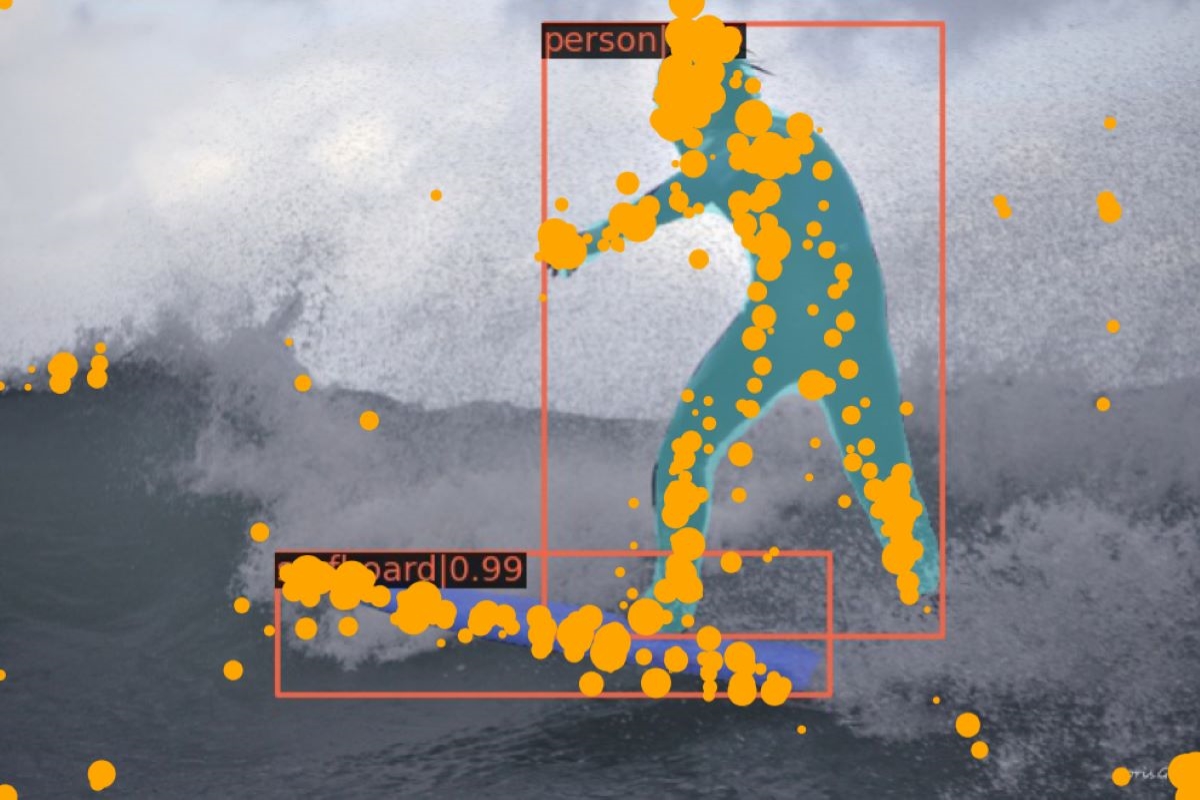}}
     \\
    \vspace{0.2pt}
    \subfloat{\label{fig:fig6_3}\includegraphics[width=0.64\linewidth]{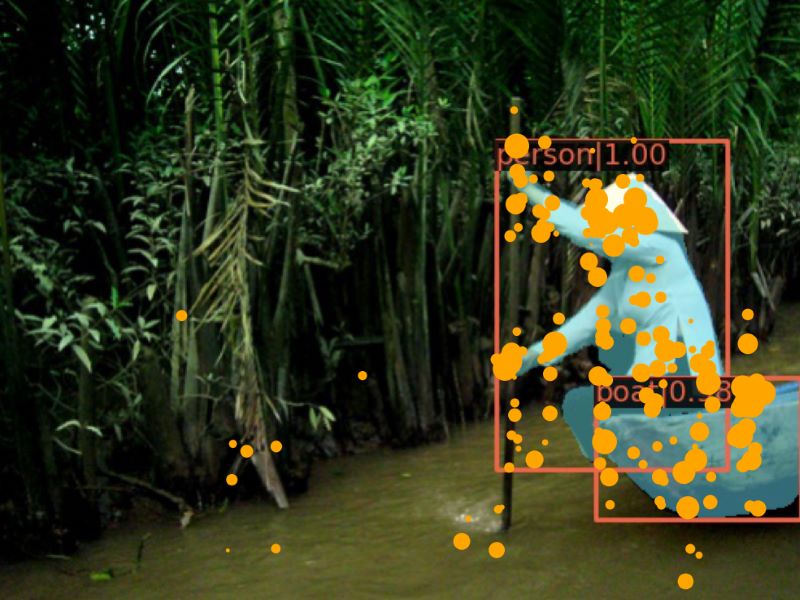}}
    \hspace{0.1pt}
    \subfloat{\label{fig:fig6_4}\includegraphics[width=0.32\linewidth]{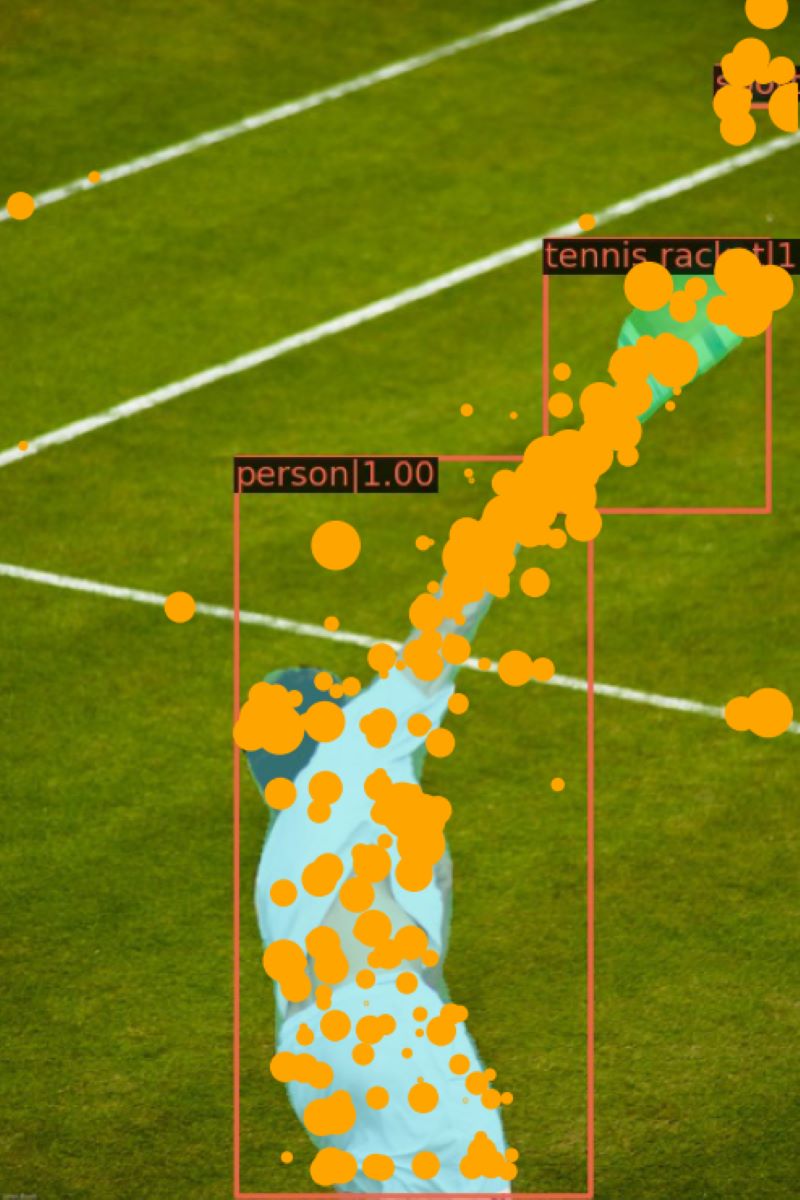}}
    \caption{Visualizations of the most important keys on COCO~\cite{coco} validation set. The orange circles show the key points with highest propagated attention scores at multiple heads. Larger radius indicate higher score. Note that the bottom right image displays a \textit{person} waving a \textit{racket} to hit a  \textit{tennis ball}.}
	\label{fig:fig6}
	\vskip -0.2in
\end{figure}

To verify the effectiveness of deformable attention, we use a similar mechanism to DCNs to visualize the most important keys across multiple deformable attention layers by propagating their attention weights. 
% Specifically, from the last deformable attention layer, we cumulatively multiply the attention weights of each deformed keys to previous layers, then average them among all queries to discover the keys with the most contributions. 
As shown in Figure~\ref{fig:fig6}, our deformable attention learns to place the keys mostly in the foreground, indicating that it focuses on the important regions of the objects, which supports our hypothesis shown in Figure~\ref{fig:fig1} of the paper. More visualizations can be found in appendix (Figure~\ref{fig:fig4},\ref{fig:figa1}).

\section{Conclusion}
% This paper presents \textit{\omlong{}}, a novel hierarchical vision Transformer leveraging deformable attention to achieve linear computational complexity with respect to input image size. \textbf{\ourmethod} outperforms Swin Transformer and other competitive baselines on both classification and dense prediction tasks as our data-dependent deformable attention is a more powerful sparse attention mechanism than data-agnostic attention patterns. 
% We believe our deformable attention module is broadly applicable to Transformer models for vision tasks.

This paper presents \textit{\omlong{}}, a novel hierarchical Vision Transformer that can be adapted to both image classification and dense prediction tasks. With deformable attention module, our model is capable of learning sparse-attention patterns in a data-dependent way and modeling geometric transformations. Extensive experiments demonstrate the effectiveness of our model over competitive baselines. We hope our work can inspire insights towards designing flexible attention techniques.

\section*{Acknowledgments}
\vspace{0.05in}

This work is supported in part by the National Science and Technology Major Project of the Ministry of Science and Technology of China under Grants 2018AAA0100701, the National Natural Science Foundation of China under Grants 61906106 and 62022048. The computational resources supporting this work are provided by Hangzhou High-Flyer AI Fundamental Research Co.,Ltd.

\vspace{0.05in}

\section*{Appendix}

\section*{A. \ourmethod{} and Deformable DETR}

In this section, we provide a detailed comparision between our proposed deformable attention and the direct adaptation from the deformable convolution \cite{DCNv1}, which is also known as the multiscale deformable attention in Deformable DETR \cite{DeformableDETR}. 

First, our deformable attention serves as a feature extractor in the vision backbones while the one in Deformable DETR 
%erran: add which
which replaces the vanilla attention in DETR \cite{detr} with a linear deformable attention, 
%erran: plays as a role of the detection head. 
plays the role of the detection head. 
Second, the $m$-th head of query $q$ in the attention in Deformable DETR with single scale is formulated as
\begin{equation}
z^{(m)}_q\!=\!\sum_{k=1}^{K}A^{(m)}_{qk}W_v\phi(x;p_q+\Delta{}p^{(m)}_{qk}),
\end{equation}
where $K$ key points are sampled from the input features, mapped by $W_v$ and then aggregated by attention weights $A^{(m)}_{qk}$. Compared to our deformable attention (Eq.(9) in the paper), this attention weights is learned from $z_q$ by a linear projection, \textit{i.e.} $A^{(m)}_{qk}=\sigma(W_\text{att}x)$, where $W_\text{att}\!\in\!\mathbb{R}^{C\times{}MK}$ is the weight matrix to predict the weights of each key on each head, after which a softmax function $\sigma$ is applied to the dimension of $K$ keys to normalize the attention score. In fact, the attention weights are predicted directly by queries instead of measuring the similarities between queries and keys. If we change the $\sigma$ function to a sigmoid, this will be a variant of modulated deformable convolution \cite{DCNv2}, hence this deformable attention is more similar to convolution rather than attention. 

Third, the deformable attention in Deformable DETR is not compatible to the dot-product attention for its enormous memory consumption mentioned in Sec.3.2 in the paper. Therefore, the linear predicted attention is used to avoid computing dot products and a smaller number of keys $K=4$ is also adopted to reduce the memory cost. 

%erran: To theoretically validate our claim, 
To experimentally validate our claim,
we replace our deformable attention modules in \ourmethod{} with the modules in \cite{DeformableDETR} to verify that the naive adaptation 
%erran: not feasible for a vision backbone. 
is inferior for vision backbone.
The comparison results are shown in Table \ref{tab:supp1}. Comparing the first and last row, we can see that under smaller memory budget, the number of keys for the deformable DETR model are set as 16 to reduce memory usage, and achieves $1.4\%$ lower performance. By comparing the third and last row, we can see that the D-DETR attention with the same number of keys as \ourmethod{} consumes 2.6$\!\times \!$ memory and 1.3$\!\times \!$ FLOPs, while the performances are still lower than \ourmethod.

\begin{table}[t]
\newcommand{\tabincell}[2]{\begin{tabular}{@{}#1@{}}#2\end{tabular}}
\begin{center}
\setlength{\tabcolsep}{0.3mm}{
\renewcommand\arraystretch{1.1}
\begin{tabular}{ccc|ccc|c}
\thickhline
\multirow{2}{*}{Attn} & Stage 3 & Stage 4 & \multirow{2}{*}{FLOPs} & \multirow{2}{*}{\#Param} & \multirow{2}{*}{Memory} & IN-1K \\
& \#Key & \#Key & & & & Acc. \\
\hline
D-DETR & 16 & 16 & 4.44G & 27.95M & 13.9GB & 80.6 \\
D-DETR & 49 & 49 & 4.83G & 31.15M & 18.8GB & 80.7 \\
D-DETR & 196 & 49 & 6.16G & 37.26M & 37.9GB & 79.2 \\
\hline
\ourmethod{} & 49 & 49 & 4.38G & 28.32M & 12.5GB & 81.8 \\
\ourmethod{} & 196 & 49 & 4.59G & 28.32M & 14.4GB & \textbf{82.0} \\
\thickhline
\end{tabular}}
\end{center}
\caption{
Comparisons of the deformable attention in \ourmethod{} with that in \cite{DeformableDETR} under different compuational budgets. The GPU memory cost is measured in a forward pass with a batch size of 64.
}
\label{tab:supp1}
\vskip -0.1in
\end{table}

% \textbf{More discussions!}

\section*{B. Adding Convolutions to \ourmethod{}}
Recent works \cite{pvtv2, cswin, rgvit, cvt} have proved that adopting convolution layers in the Vision Transformer architecture can further improve model performances. For example, using convolutional patch embedding can generally boost model performances by $0.5\%\sim 1.0\%$ on ImageNet classification tasks. It is worth noticing that our proposed \ourmethod{} can readily combine with these techniques, while we maintain the convolution-free architecture in the main paper to perform fair comparison with baselines.

To fully explore the capacity of \ourmethod{}, we substitute the patch embedding layers in the original model with strided and overlapped convolutions. The comparison results are shown in Table \ref{tab:supp2}, where baseline models have similar modifications. It is shown that our model with additional convolution modules achieve $0.7\%$ improvement comparing to the original version, and consistently outperform other baselines.
% Many recent works \cite{pvtv2, cswin, rgvit, cvt} adopt convolutions into Vision Transformers, either using convolutinal patch embedding or inserting depthwise convolutions in the FFN of Transformer block. We also provide some variants of our \ourmethod{} to further improve performances, as shown in \ref{tab:supp2}, \ourmethod{} achieves competitive results among recent works.

\begin{table}[t]
\newcommand{\tabincell}[2]{\begin{tabular}{@{}#1@{}}#2\end{tabular}}
\begin{center}
\setlength{\tabcolsep}{1.0mm}{
\renewcommand\arraystretch{1.1}
\begin{tabular}{l|cc|c}
\thickhline
\multicolumn{4}{c}{\textbf{ImageNet-1K Classification}} \\
Method & FLOPs & \#Param & Top-1 Acc.\\
\hline
% Twins-PCPVT-S \cite{twins} & 3.8G & 24.1M & 81.2 \\
% CPVT-S-GAP \cite{cpvt} & - & 23M & 81.5 \\
CvT-13 \cite{cvt} & 4.5G & 20M & 81.6 \\
CoAt-Lite Small \cite{coat} & 4.0G & 20M & 81.9 \\
CeiT-S \cite{ceit} & 4.8G & 24M & 82.0 \\
PVTv2-B2 \cite{pvtv2} & 4.0G & 25M & 82.0 \\
CoAt Small \cite{coat} & 12.6G & 22M & 82.1 \\
RegionViT-S \cite{rgvit} & 5.3G & 31M & 82.5 \\
% CvT-21 \cite{cvt} & 7.1G & 32M & 82.5 \\
% CSWin-T \cite{cswin} & 4.3G & 23M & 82.7 \\
\hline
\ourmethod-T & 4.6G & 28M & 82.0 \\
\rowcolor{mygray}
\textbf{\ourmethod-T*} & 4.8G & 30M & \textbf{82.7} \\
\thickhline
\end{tabular}}
\end{center}
\caption{
%Comparisons of FLOPS and parameters against accuracy on ImageNet-1K classification task.
Comparisons of \ourmethod{} with other vision transformer backbones on FLOPS, parameters, accuracy on the ImageNet-1K classification task. \ourmethod-T refers to the original version. \ourmethod-T* refers to the model with convolutional patch embeddings.
}
\label{tab:supp2}
\vskip -0.15in
\end{table}

\begin{figure}
	\centering
	\subfloat{\label{fig:fig4_1}\includegraphics[width=0.20\linewidth]{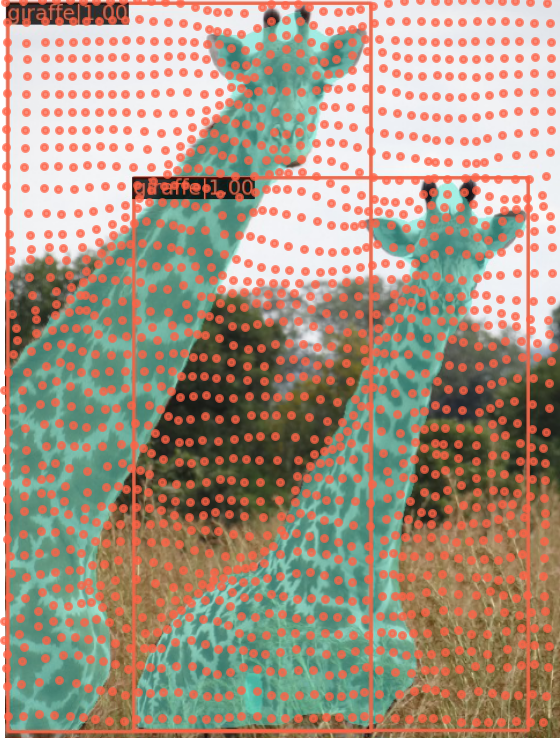}} \hspace{1pt}
	\subfloat{\label{fig:fig4_2}\includegraphics[width=0.40\linewidth]{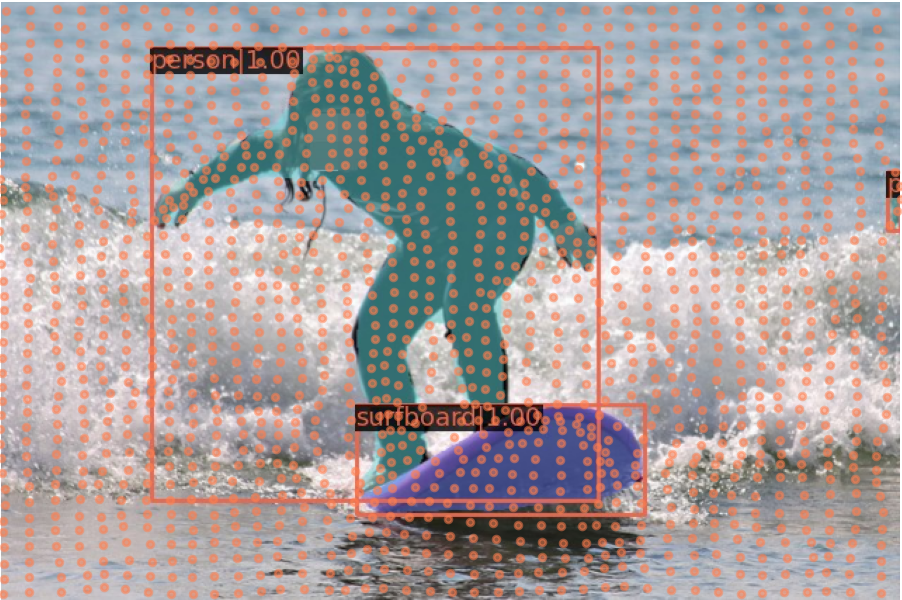}} \hspace{1pt}
	\subfloat{\label{fig:fig4_3}\includegraphics[width=0.355\linewidth]{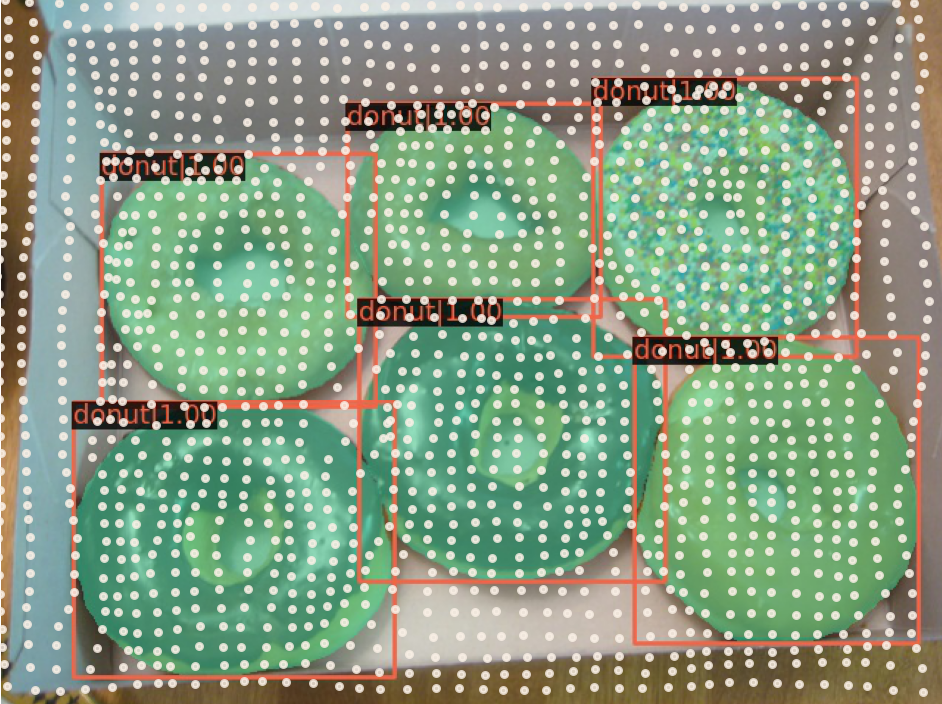}} \\
	\vspace{1pt}
	\subfloat{\label{fig:fig4_4}\includegraphics[width=0.20\linewidth]{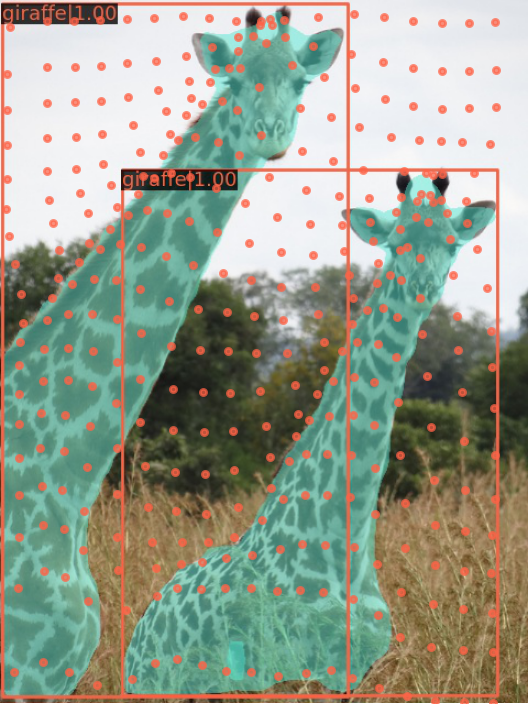}} \hspace{1pt}
	\subfloat{\label{fig:fig4_5}\includegraphics[width=0.40\linewidth]{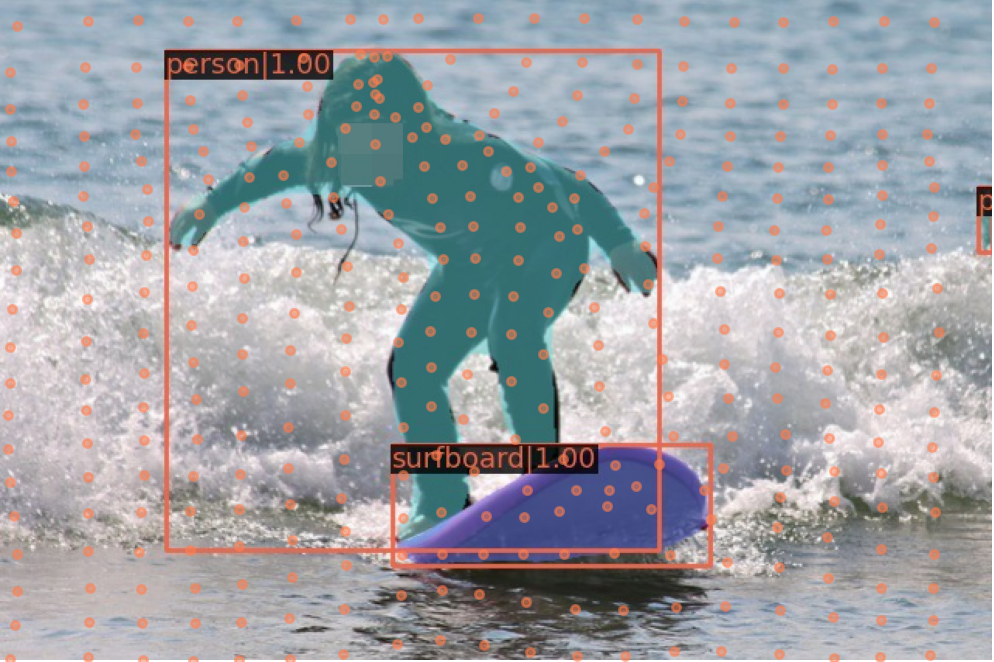}} \hspace{1pt}
	\subfloat{\label{fig:fig4_6}\includegraphics[width=0.355\linewidth]{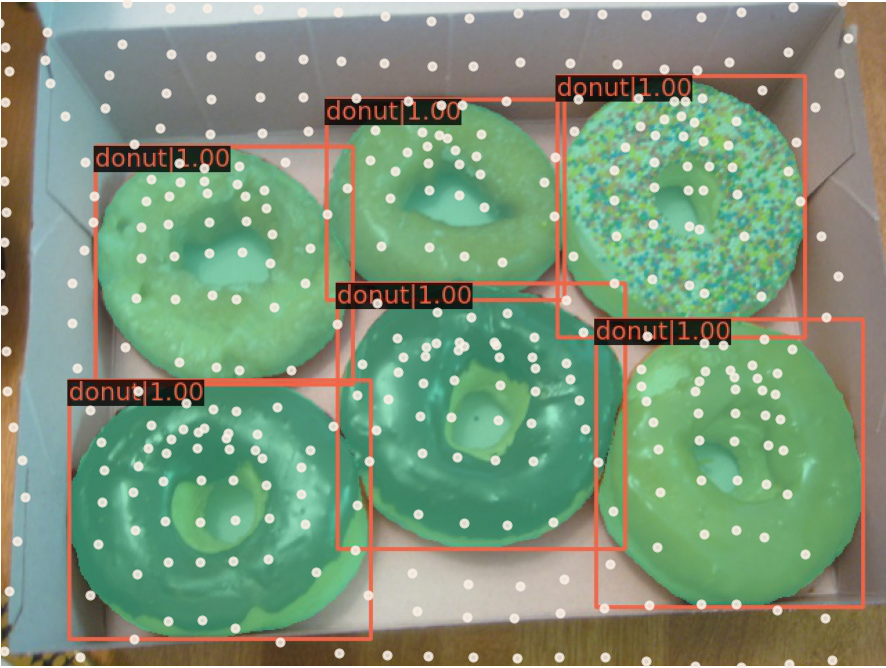}} \\
	
	\caption{Visualizations on COCO \cite{coco} of learned sampling locations in deformable attention at Stage 3 (first row) and Stage 4 (second row) of \ourmethod{}. The orange and yellow points show one group of deformed points. The detection bounding boxes and segmentation masks are also presented to indicate the targets.}
	\label{fig:fig4}
	\vskip -0.15in
\end{figure}

\begin{figure}
	\centering
	\subfloat{\label{fig:figa1_1}\includegraphics[width=0.35\linewidth]{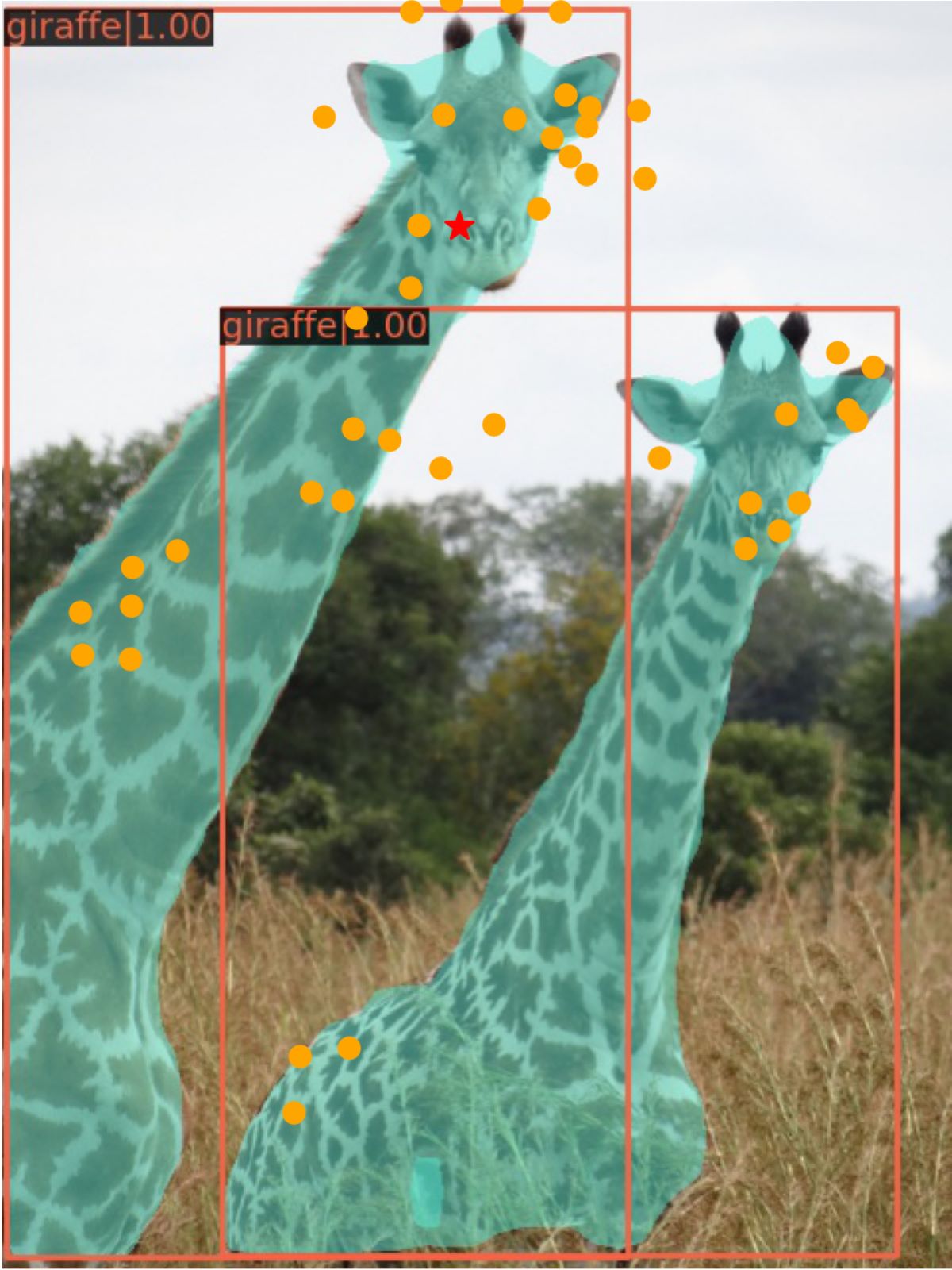}} \hspace{1pt}
	\subfloat{\label{fig:figa1_2}\includegraphics[width=0.61\linewidth]{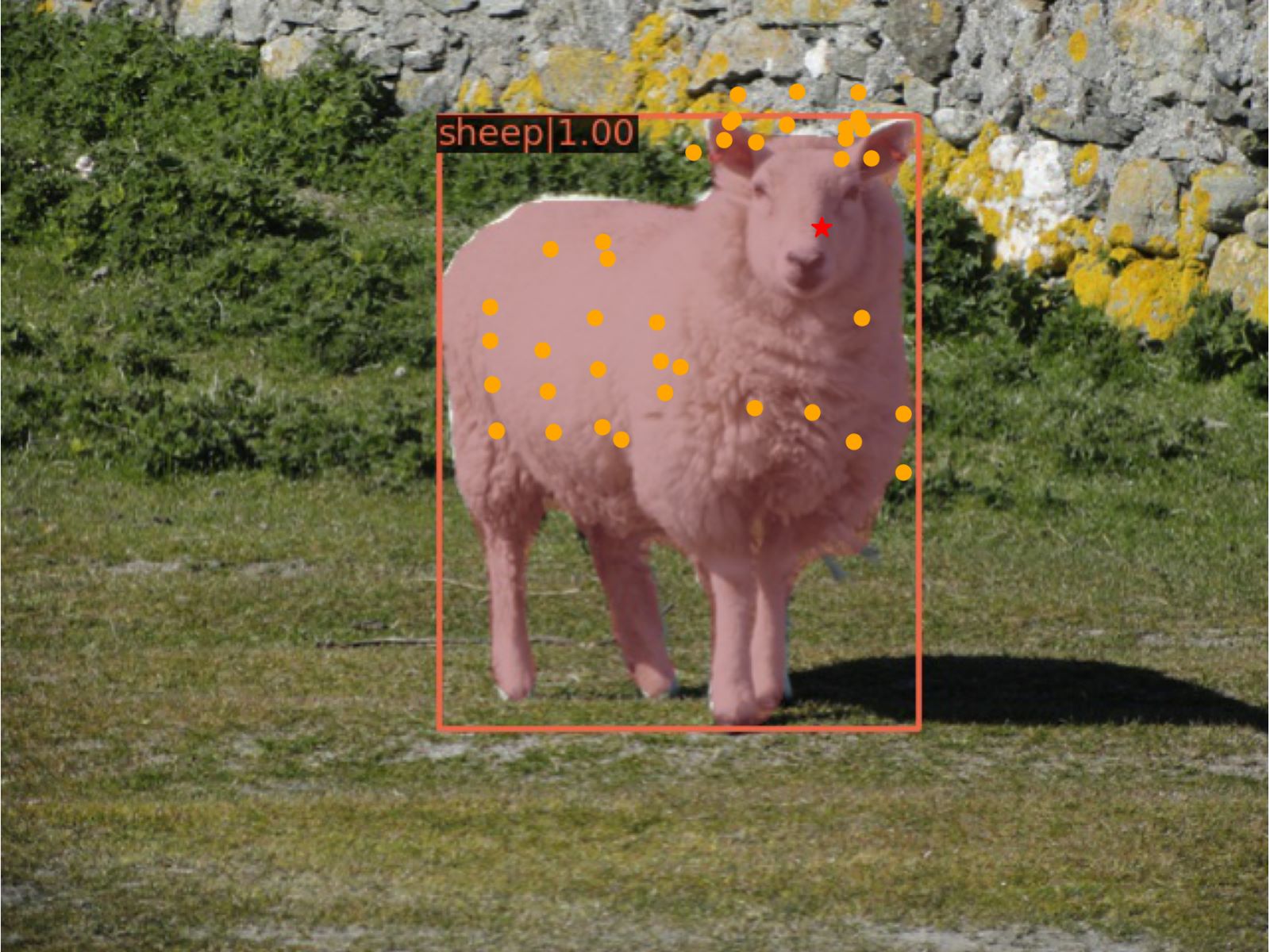}}
	\\
	\vspace{1pt}
	\subfloat{\label{fig:figa1_5}\includegraphics[width=0.35\linewidth]{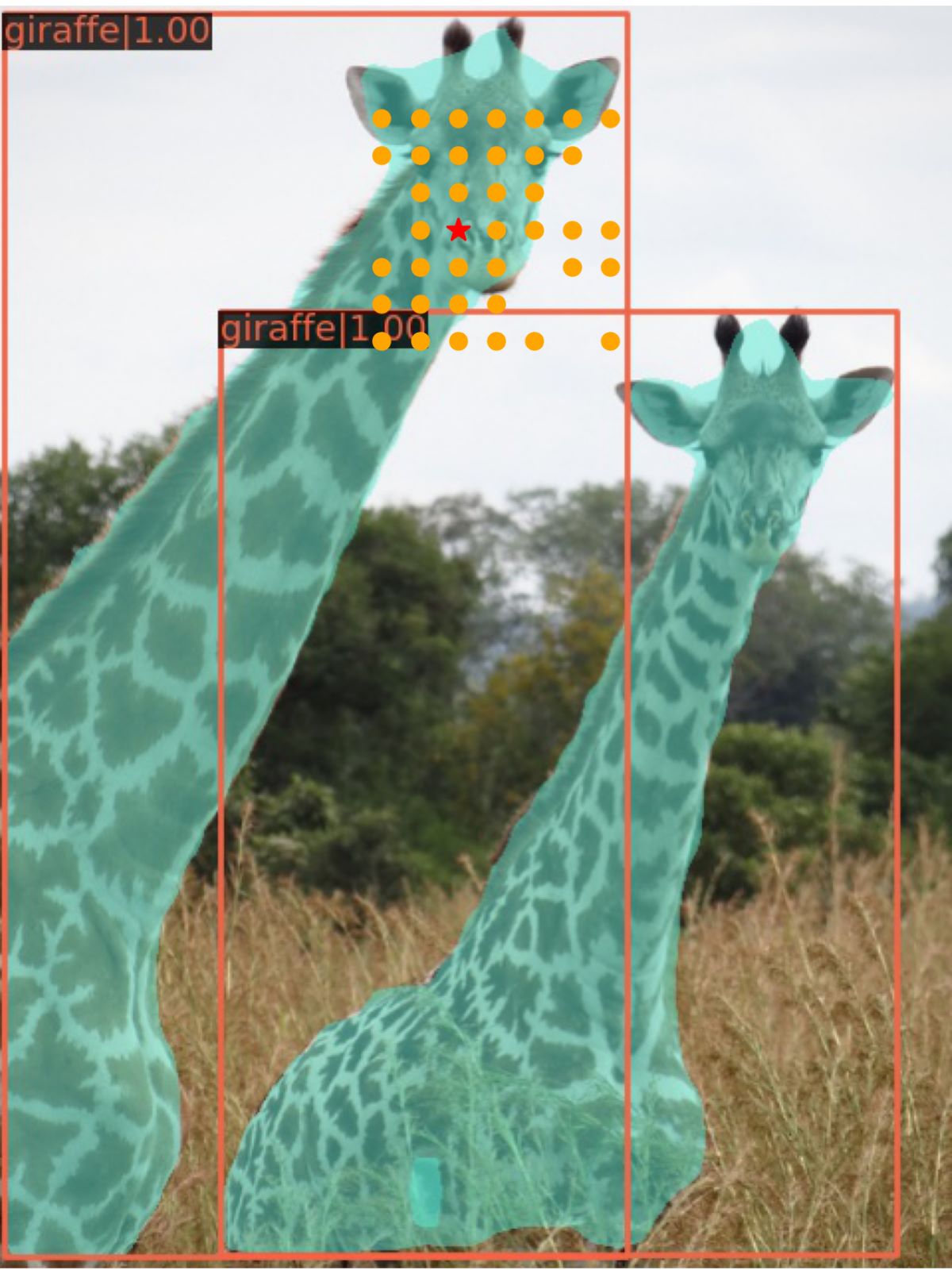}} \hspace{1pt}
	\subfloat{\label{fig:figa1_6}\includegraphics[width=0.61\linewidth]{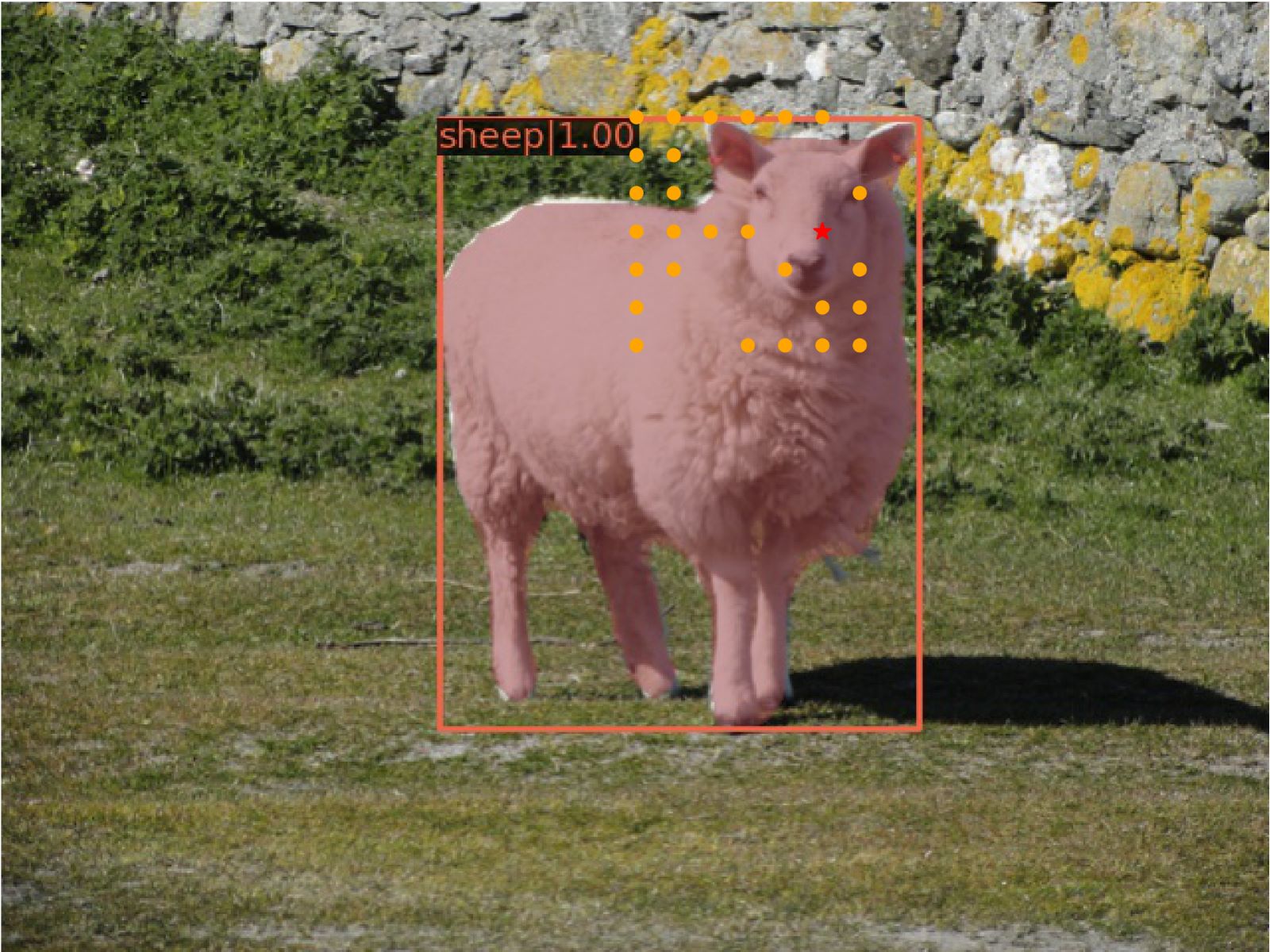}}
	\\
	\vspace{1pt}
	\subfloat{\label{fig:figa1_3}\includegraphics[width=0.48\linewidth]{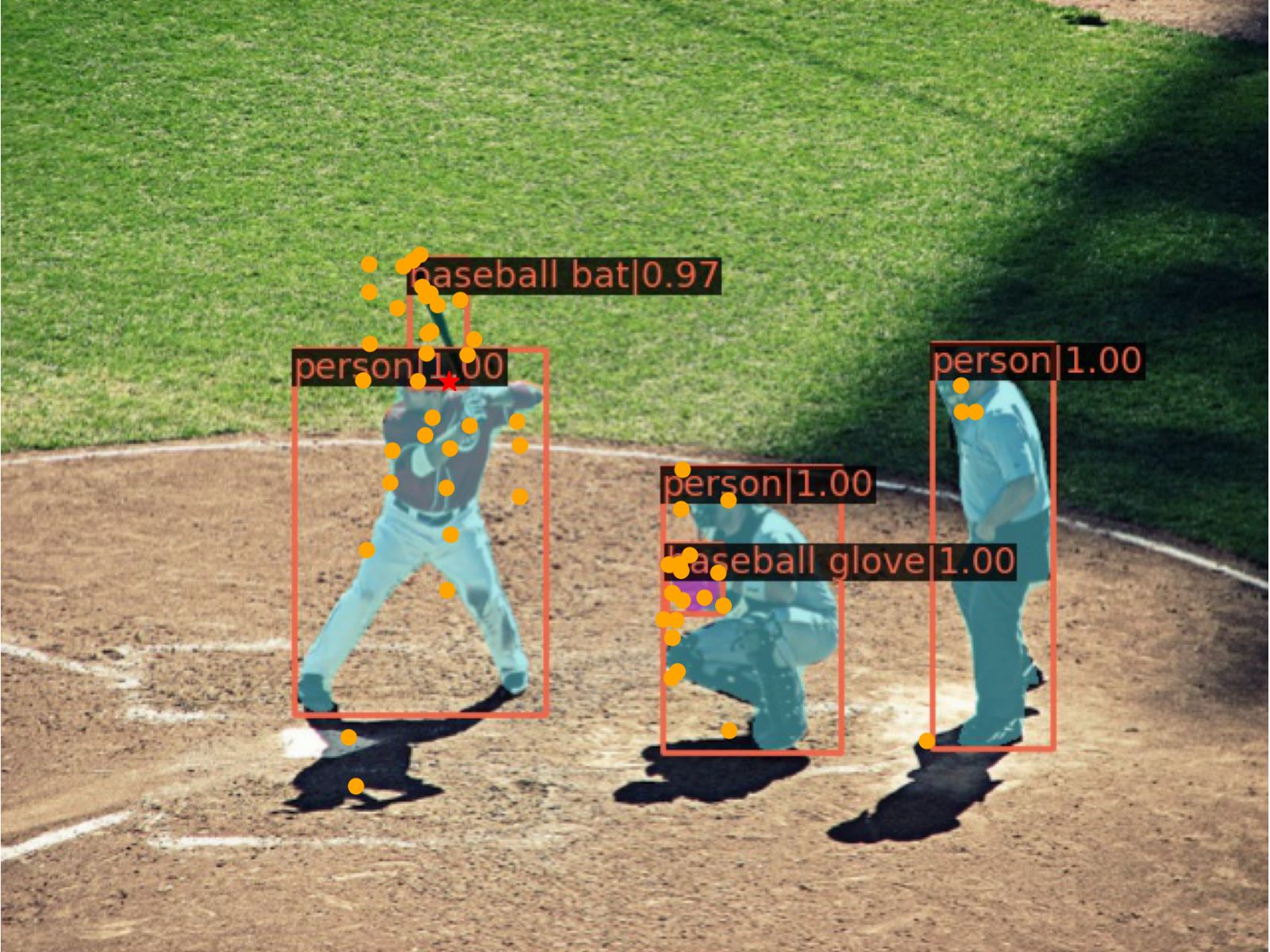}} \hspace{1pt} 
	\subfloat{\label{fig:figa1_4}\includegraphics[width=0.48\linewidth]{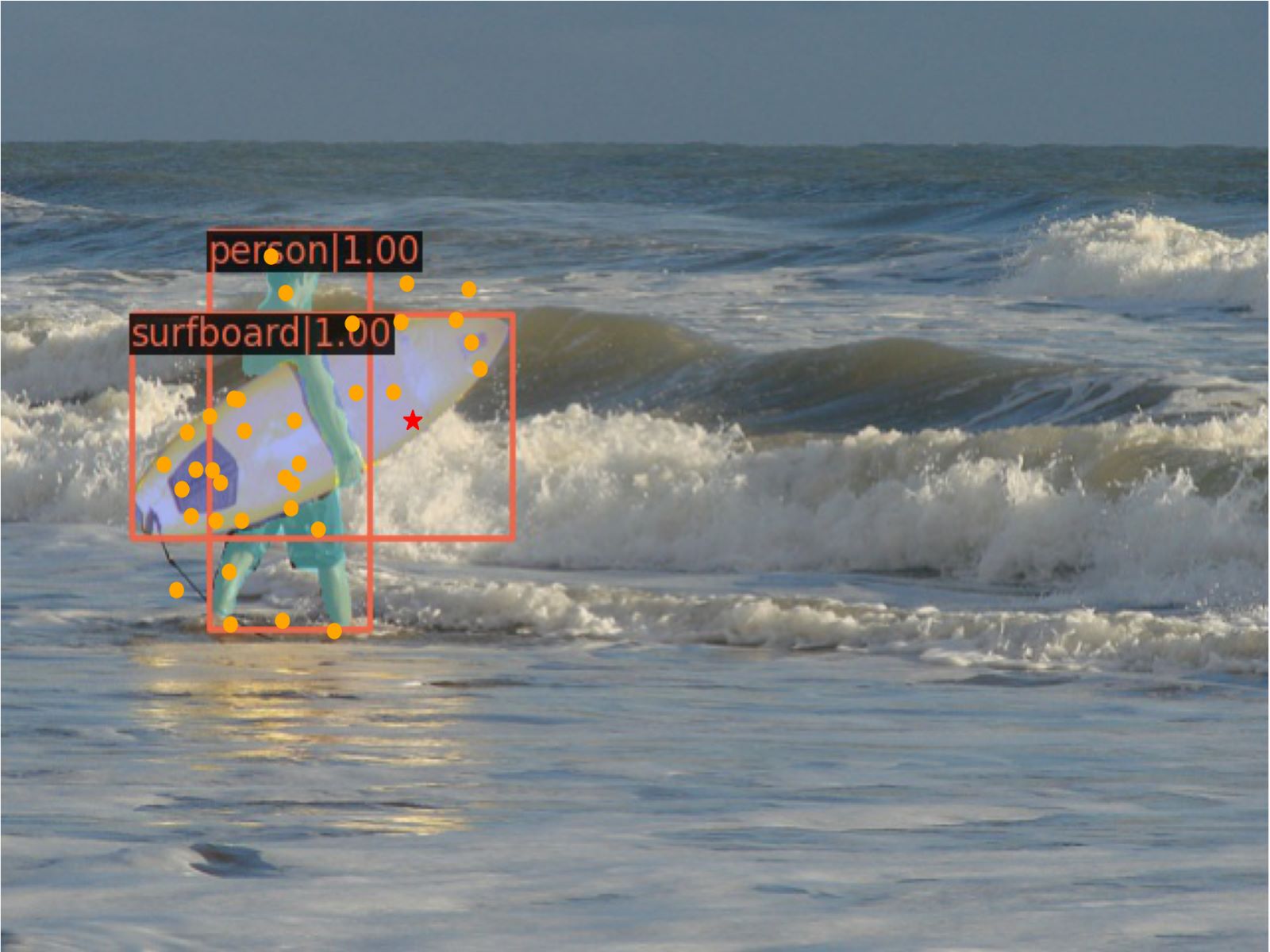}} 
	\\
	\vspace{1pt} 
	\subfloat{\label{fig:figa1_7}\includegraphics[width=0.48\linewidth]{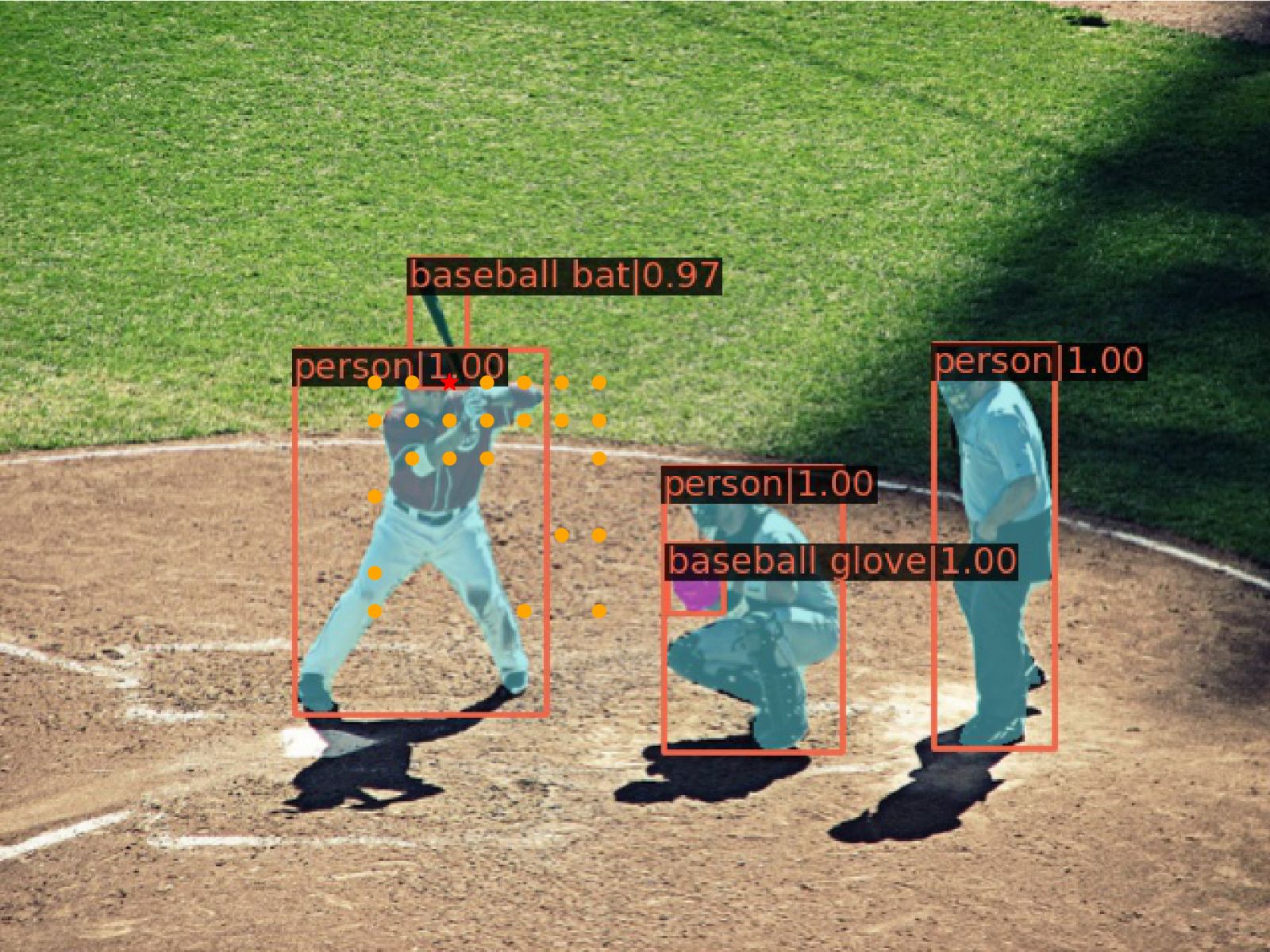}} \hspace{1pt}
	\subfloat{\label{fig:figa1_8}\includegraphics[width=0.48\linewidth]{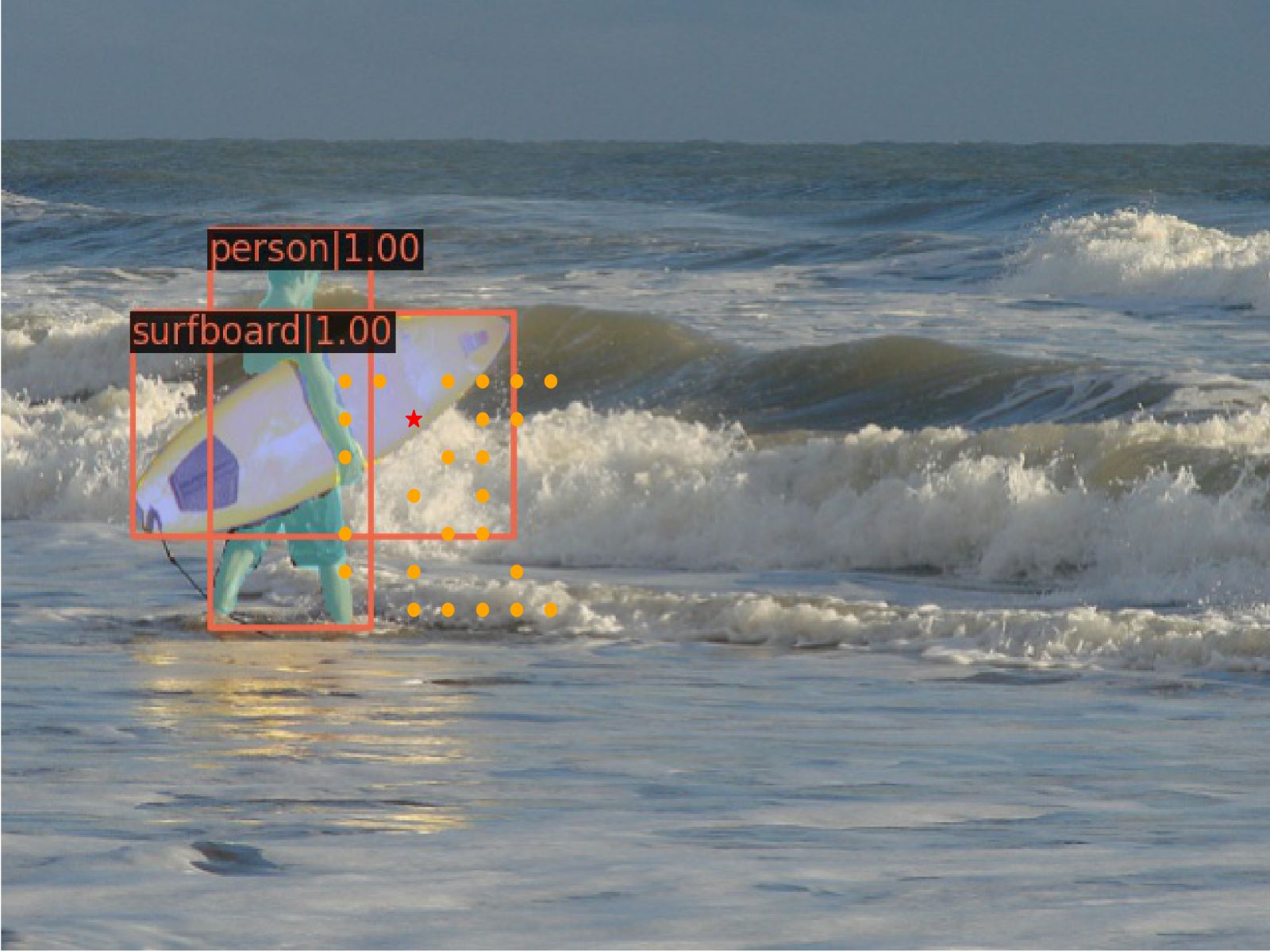}}
	\caption{Visualizations on COCO \cite{coco} validation set. The red star denotes a query point, the orange dots are the keys with higher attention scores in the last layer. The images in the first and third rows depict our \ourmethod{} attention and Swin Transformers' \cite{Swin} are shown in the second and fourth rows. The detection bounding boxes and segmentation masks are also presented to indicate the targets.}
	\label{fig:figa1}
	\vskip -0.15in
\end{figure}

% \textbf{Not done yet. Maybe remove this section.}

\section*{C. More Visualizations}

We visualize examples of learned deformed locations in our \ourmethod{} to verify the effectiveness of our method. As illustrated in Figure~\ref{fig:fig4}, the sampling points are depicted on the top of the object detection boxes and instance segmentation masks, from which we can see that the points are shifted to the target objects. In the left column, the deformed points are contracted to two target giraffes, while other points are keeping a nearly uniform grid with small offsets. In the middle column, the deformed points distribute densely among the person's body and the surfing board both in the two stages. The right column shows the deformed points focus well to each of the six donuts, which shows our model has the ability to better model geometric shapes even with multiple targets. The above visualizations demonstrate that \ourmethod{} learns meaningful offsets to sample better keys for attention to improve the performances on various vision tasks.

We also provide visualization results of the attention map given specific query tokens, and compare with Swin Transformer \cite{Swin} in Figure \ref{fig:figa1}. We show key tokens with the highest attention values. It can be observed that our model focus on the more relevent part. As a showcase, our model allocates most attention to foreground objects, \textit{e.g.}, both gireffas in the first row. On the other hand, the region of interests in Swin Transformer is comparably local and fail to distinguish foreground from background, which is depicted in the last surfboard.

{\small
\bibliographystyle{ieee_fullname}
\bibliography{egbib}
}
\end{document}